\documentclass[letterpaper]{article}
\usepackage[preprint]{aaai2027}
\usepackage[hyphens]{url}
\usepackage{graphicx}
\usepackage{natbib}
\usepackage{caption}
\usepackage{amsmath}
\usepackage{amssymb}
\usepackage{algorithm}
\usepackage{algorithmic}
\usepackage{booktabs}

\chardef\OtherCatcode=12
\newcommand{\ProtocolBenchmarkTaskCount}{300}
\newcommand{\ProtocolChallengeTaskCount}{64}
\newcommand{\ProtocolInitialRound}{1}
\newcommand{\ProtocolInitialPromptCandidateCount}{1}

\newcommand{\ProtocolMaxLLMCalls}{6}
\newcommand{\ProtocolLaterPromptCandidateCount}{3}
\newcommand{\ProtocolGoldTightASTMargin}{10}
\newcommand{\ProtocolGoldLooseASTMargin}{25}

\newcommand{\NoSymbolicPairCount}{9}
\newcommand{\NoSymbolicWinCount}{9}

\newcommand{\NoSymbolicMinDeltaPoints}{+6.2}
\newcommand{\NoSymbolicMaxDeltaPoints}{+25.0}
\newcommand{\NoSymbolicCIExcludesZeroCount}{7}
\newcommand{\NoSymbolicSamplingProgressCount}{9}
\newcommand{\NoSymbolicLowerMeanCallsCount}{9}

\newcommand{\FrontierConfigCount}{7}

\newcommand{\RepairReducedMinPercent}{83.1}
\newcommand{\RepairReducedMaxPercent}{98.4}
\newcommand{\RepairExactMinPercent}{1.2}
\newcommand{\RepairExactMaxPercent}{7.8}

\newcommand{\GoldOrSmallerTaskExactMin}{100.0}
\newcommand{\GoldBloatTaskExactMax}{1.7}

\newcommand{\CompactAllN_B}{185}
\newcommand{\CompactAllDelta_B}{+19.9}
\newcommand{\CompactAllLo_B}{+13.3}
\newcommand{\CompactAllHi_B}{+27.0}
\newcommand{\CompactNoGoldN_B}{159}
\newcommand{\CompactNoGoldDelta_B}{+8.2}
\newcommand{\CompactNoGoldLo_B}{+3.3}
\newcommand{\CompactNoGoldHi_B}{+13.8}
\newcommand{\CompactAboveGoldN_B}{136}
\newcommand{\CompactAboveGoldDelta_B}{-1.5}
\newcommand{\CompactAboveGoldLo_B}{-4.3}
\newcommand{\CompactAboveGoldHi_B}{+1.5}
\newcommand{\CompactAllN_C}{64}
\newcommand{\CompactAllDelta_C}{+7.3}
\newcommand{\CompactAllLo_C}{0.0}
\newcommand{\CompactAllHi_C}{+17.2}
\newcommand{\CompactNoGoldN_C}{61}
\newcommand{\CompactNoGoldDelta_C}{+3.1}
\newcommand{\CompactNoGoldLo_C}{-2.9}
\newcommand{\CompactNoGoldHi_C}{+11.3}
\newcommand{\CompactAboveGoldN_C}{55}
\newcommand{\CompactAboveGoldDelta_C}{-3.3}
\newcommand{\CompactAboveGoldLo_C}{-7.1}
\newcommand{\CompactAboveGoldHi_C}{+1.6}

\newcommand{\FrontierInitialMeanBenchmarkPercent}{4.7}
\newcommand{\FrontierFinalMeanBenchmarkPercent}{29.0}
\newcommand{\FrontierInitialMeanChallengePercent}{29.9}
\newcommand{\FrontierFinalMeanChallengePercent}{59.4}

\newcommand{\ZThreePrenexBenchmarkValidPercent}{41.3}

\newcommand{\ZThreePrenexChallengeValidPercent}{15.6}

\newcommand{\ZThreeAdMixBenchmarkValidPercent}{59.7}

\newcommand{\ZThreeAdMixChallengeValidPercent}{28.1}
\newcommand{\SymbolicBaselineTimeoutMinutes}{30}

\newcommand{\ZThreePrenexBenchmarkAccTwentyFivePercent}{33.0}

\newcommand{\ZThreePrenexBenchmarkHExactPercent}{14.0}
\newcommand{\ZThreePrenexBenchmarkASTMedian}{33.0}
\newcommand{\ZThreePrenexBenchmarkBothCount}{227}
\newcommand{\ZThreePrenexBenchmarkHFOnlyCount}{171}
\newcommand{\ZThreePrenexBenchmarkZThreeOnlyCount}{21}
\newcommand{\ZThreePrenexBenchmarkUnionCount}{419}

\newcommand{\ZThreePrenexChallengeAccTwentyFivePercent}{10.9}

\newcommand{\ZThreePrenexChallengeHExactPercent}{4.7}
\newcommand{\ZThreePrenexChallengeASTMedian}{35.5}
\newcommand{\ZThreePrenexChallengeBothCount}{43}
\newcommand{\ZThreePrenexChallengeHFOnlyCount}{139}
\newcommand{\ZThreePrenexChallengeZThreeOnlyCount}{7}
\newcommand{\ZThreePrenexChallengeUnionCount}{189}

\newcommand{\ZThreeAdMixBenchmarkAccTwentyFivePercent}{58.3}

\newcommand{\ZThreeAdMixBenchmarkHExactPercent}{54.3}
\newcommand{\ZThreeAdMixBenchmarkASTMedian}{16.0}
\newcommand{\ZThreeAdMixBenchmarkBothCount}{289}
\newcommand{\ZThreeAdMixBenchmarkHFOnlyCount}{109}
\newcommand{\ZThreeAdMixBenchmarkZThreeOnlyCount}{69}
\newcommand{\ZThreeAdMixBenchmarkUnionCount}{467}

\newcommand{\ZThreeAdMixChallengeAccTwentyFivePercent}{28.1}

\newcommand{\ZThreeAdMixChallengeHExactPercent}{25.0}
\newcommand{\ZThreeAdMixChallengeASTMedian}{16.0}
\newcommand{\ZThreeAdMixChallengeBothCount}{67}
\newcommand{\ZThreeAdMixChallengeHFOnlyCount}{115}
\newcommand{\ZThreeAdMixChallengeZThreeOnlyCount}{23}
\newcommand{\ZThreeAdMixChallengeUnionCount}{205}

\newcommand{\SymbolicCascadeBenchmarkPrenexHFFinalValidPercent}{69.8}
\newcommand{\SymbolicCascadeBenchmarkPrenexHFResidualValidPercent}{48.6}
\newcommand{\SymbolicCascadeBenchmarkPrenexHFCallsPerTask}{2.34}
\newcommand{\SymbolicCascadeBenchmarkPrenexHFCallsSavedPercent}{25.5}
\newcommand{\SymbolicCascadeBenchmarkPrenexRepeatFinalValidPercent}{61.2}
\newcommand{\SymbolicCascadeBenchmarkPrenexRepeatResidualValidPercent}{33.8}
\newcommand{\SymbolicCascadeBenchmarkPrenexRepeatCallsPerTask}{2.77}
\newcommand{\SymbolicCascadeBenchmarkPrenexRepeatCallsSavedPercent}{34.0}
\newcommand{\SymbolicCascadeBenchmarkAdMixHFFinalValidPercent}{77.8}
\newcommand{\SymbolicCascadeBenchmarkAdMixHFResidualValidPercent}{45.0}
\newcommand{\SymbolicCascadeBenchmarkAdMixHFCallsPerTask}{1.68}
\newcommand{\SymbolicCascadeBenchmarkAdMixHFCallsSavedPercent}{46.6}
\newcommand{\SymbolicCascadeBenchmarkAdMixRepeatFinalValidPercent}{69.5}
\newcommand{\SymbolicCascadeBenchmarkAdMixRepeatResidualValidPercent}{24.4}
\newcommand{\SymbolicCascadeBenchmarkAdMixRepeatCallsPerTask}{2.08}
\newcommand{\SymbolicCascadeBenchmarkAdMixRepeatCallsSavedPercent}{50.6}
\newcommand{\SymbolicCascadeChallengePrenexHFFinalValidPercent}{59.1}
\newcommand{\SymbolicCascadeChallengePrenexHFResidualValidPercent}{51.5}
\newcommand{\SymbolicCascadeChallengePrenexHFCallsPerTask}{3.38}
\newcommand{\SymbolicCascadeChallengePrenexHFCallsSavedPercent}{8.3}
\newcommand{\SymbolicCascadeChallengePrenexRepeatFinalValidPercent}{47.8}
\newcommand{\SymbolicCascadeChallengePrenexRepeatResidualValidPercent}{38.1}
\newcommand{\SymbolicCascadeChallengePrenexRepeatCallsPerTask}{3.75}
\newcommand{\SymbolicCascadeChallengePrenexRepeatCallsSavedPercent}{10.9}
\newcommand{\SymbolicCascadeChallengeAdMixHFFinalValidPercent}{64.1}
\newcommand{\SymbolicCascadeChallengeAdMixHFResidualValidPercent}{50.0}
\newcommand{\SymbolicCascadeChallengeAdMixHFCallsPerTask}{2.93}
\newcommand{\SymbolicCascadeChallengeAdMixHFCallsSavedPercent}{20.5}
\newcommand{\SymbolicCascadeChallengeAdMixRepeatFinalValidPercent}{54.1}
\newcommand{\SymbolicCascadeChallengeAdMixRepeatResidualValidPercent}{36.1}
\newcommand{\SymbolicCascadeChallengeAdMixRepeatCallsPerTask}{3.34}
\newcommand{\SymbolicCascadeChallengeAdMixRepeatCallsSavedPercent}{20.7}
\newcommand{\CascadePrenexHFBThreeZeroZeroDeepseekOnePromptVsRepeatValidPercent}{71.3}
\newcommand{\CascadePrenexHFBThreeZeroZeroDeepseekOnePromptVsRepeatHExactPercent}{30.7}
\newcommand{\CascadePrenexHFBThreeZeroZeroDeepseekOnePromptVsRepeatCallsPerTask}{2.35}
\newcommand{\CascadePrenexHFBThreeZeroZeroDeepseekOnePromptVsRepeatCallsSavedPercent}{26.4}
\newcommand{\CascadePrenexHFBThreeZeroZeroGeminiOnePromptVsRepeatValidPercent}{68.3}
\newcommand{\CascadePrenexHFBThreeZeroZeroGeminiOnePromptVsRepeatHExactPercent}{36.7}
\newcommand{\CascadePrenexHFBThreeZeroZeroGeminiOnePromptVsRepeatCallsPerTask}{2.33}
\newcommand{\CascadePrenexHFBThreeZeroZeroGeminiOnePromptVsRepeatCallsSavedPercent}{24.6}
\newcommand{\CascadePrenexHFCSixFourNativeSolOnePromptVsRepeatValidPercent}{89.1}
\newcommand{\CascadePrenexHFCSixFourNativeSolOnePromptVsRepeatHExactPercent}{29.7}
\newcommand{\CascadePrenexHFCSixFourNativeSolOnePromptVsRepeatCallsPerTask}{1.94}
\newcommand{\CascadePrenexHFCSixFourNativeSolOnePromptVsRepeatCallsSavedPercent}{7.5}
\newcommand{\CascadePrenexHFCSixFourNativeTerraOnePromptVsRepeatValidPercent}{81.2}
\newcommand{\CascadePrenexHFCSixFourNativeTerraOnePromptVsRepeatHExactPercent}{29.7}
\newcommand{\CascadePrenexHFCSixFourNativeTerraOnePromptVsRepeatCallsPerTask}{2.61}
\newcommand{\CascadePrenexHFCSixFourNativeTerraOnePromptVsRepeatCallsSavedPercent}{6.2}
\newcommand{\CascadePrenexHFCSixFourNativeGrokFourFiveOnePromptVsRepeatValidPercent}{53.1}
\newcommand{\CascadePrenexHFCSixFourNativeGrokFourFiveOnePromptVsRepeatHExactPercent}{17.2}
\newcommand{\CascadePrenexHFCSixFourNativeGrokFourFiveOnePromptVsRepeatCallsPerTask}{4.11}
\newcommand{\CascadePrenexHFCSixFourNativeGrokFourFiveOnePromptVsRepeatCallsSavedPercent}{7.1}

\newcommand{\CascadePrenexHFBThreeZeroZeroKimiTwoSevenValidPercent}{68.7}
\newcommand{\CascadePrenexHFBThreeZeroZeroKimiTwoSevenHExactPercent}{33.0}
\newcommand{\CascadePrenexHFBThreeZeroZeroKimiTwoSevenCallsPerTask}{2.40}
\newcommand{\CascadePrenexHFBThreeZeroZeroKimiTwoSevenCallsSavedPercent}{24.2}
\newcommand{\CascadePrenexHFCSixFourFableFiveValidPercent}{76.6}
\newcommand{\CascadePrenexHFCSixFourFableFiveHExactPercent}{25.0}
\newcommand{\CascadePrenexHFCSixFourFableFiveCallsPerTask}{2.20}
\newcommand{\CascadePrenexHFCSixFourFableFiveCallsSavedPercent}{6.6}
\newcommand{\CascadePrenexHFCSixFourLunaValidPercent}{50.0}
\newcommand{\CascadePrenexHFCSixFourLunaHExactPercent}{20.3}
\newcommand{\CascadePrenexHFCSixFourLunaCallsPerTask}{3.73}
\newcommand{\CascadePrenexHFCSixFourLunaCallsSavedPercent}{7.7}

\newcommand{\CascadeComparisonCount}{4}

\newcommand{\CascadeHFStandaloneGainMinPoints}{2.2}
\newcommand{\CascadeHFStandaloneGainMaxPoints}{11.5}

\newcommand{\CascadeResidualGainMinPoints}{13.3}
\newcommand{\CascadeResidualGainMaxPoints}{20.7}
\newcommand{\CascadeHFCallSavingsMinPercent}{8.3}
\newcommand{\CascadeHFCallSavingsMaxPercent}{46.6}

\newcommand{\PairedQualityConfigCount}{7}

\newcommand{\PrenexCascadeConfigCount}{8}
\newcommand{\SimplificationBenchmarkInputs}{1330}
\newcommand{\SimplificationBenchmarkSmallerCount}{461}
\newcommand{\SimplificationBenchmarkSmallerPercent}{34.7}
\newcommand{\SimplificationBenchmarkASTPreMean}{40.9}
\newcommand{\SimplificationBenchmarkASTPostMean}{32.5}
\newcommand{\SimplificationBenchmarkHExactPrePercent}{30.4}
\newcommand{\SimplificationBenchmarkHExactPostPercent}{33.6}

\newcommand{\SimplificationChallengeInputs}{379}
\newcommand{\SimplificationChallengeSmallerCount}{173}
\newcommand{\SimplificationChallengeSmallerPercent}{45.6}
\newcommand{\SimplificationChallengeASTPreMean}{128.3}
\newcommand{\SimplificationChallengeASTPostMean}{98.4}
\newcommand{\SimplificationChallengeHExactPrePercent}{18.9}
\newcommand{\SimplificationChallengeHExactPostPercent}{20.0}

\newcommand{\SymbolicCascadeBenchmarkPrenexHFHExactPercent}{33.7}
\newcommand{\SymbolicCascadeBenchmarkPrenexRepeatHExactPercent}{32.0}
\newcommand{\SymbolicCascadeBenchmarkAdMixHFHExactPercent}{61.8}
\newcommand{\SymbolicCascadeBenchmarkAdMixRepeatHExactPercent}{59.7}
\newcommand{\SymbolicCascadeChallengePrenexHFHExactPercent}{23.4}
\newcommand{\SymbolicCascadeChallengePrenexRepeatHExactPercent}{23.4}
\newcommand{\SymbolicCascadeChallengeAdMixHFHExactPercent}{34.1}
\newcommand{\SymbolicCascadeChallengeAdMixRepeatHExactPercent}{34.4}
\newcommand{\QualityBenchmarkValidPercent}{29.0}
\newcommand{\QualityBenchmarkHExactPercent}{17.3}
\newcommand{\QualityChallengeValidPercent}{59.4}
\newcommand{\QualityChallengeHExactPercent}{23.2}
\newcommand{\TwoFrontEndCascadeConfigCount}{7}
\newcommand{\PosthocMatchedPairCount}{9}
\newcommand{\PosthocHExactWinCount}{8}
\newcommand{\PosthocAccGoldTwentyFiveWinCount}{8}
\newcommand{\PosthocBenchmarkPairCount}{3}
\newcommand{\PosthocBenchmarkAccGainMinPoints}{+7.7}
\newcommand{\PosthocBenchmarkAccGainMaxPoints}{+9.3}
\newcommand{\PosthocBenchmarkAccCIExcludesZeroCount}{3}
\newcommand{\PosthocBootstrapConfidencePercent}{95}
\newcommand{\PosthocRepairSelectedCount}{58}
\newcommand{\PosthocRepairInvalidParentCount}{40}
\newcommand{\PosthocRepairInvalidParentPercent}{69}
\newcommand{\TableOneAccGoldTwentyFiveCelleZeroFourSeveneThreeOneSixThreeadaThreeaEighta}{45.0}
\newcommand{\TableOneAccGoldTwentyFiveCellZeroTwoFivedcdFourfaebOneSevenEightae}{37.3}
\newcommand{\TableOneAccGoldTwentyFiveCelldSixTwocSixSixSixTwoSevenFourTwoZeroThreeZeroThreeFour}{54.7}
\newcommand{\TableOneAccGoldTwentyFiveCellaEightZerobcEightOnefOneOneNineSevenbThreeThreeNine}{45.3}
\newcommand{\TableOneAccGoldTwentyFiveCellThreeSevenFiveFivecFourOneNineZeroFourTwoFourdZeroSixd}{46.7}
\newcommand{\TableOneAccGoldTwentyFiveCellNineeTwodNineaTwoTwoadaThreeOneSevenSevenSix}{39.0}
\newcommand{\TableOneAccGoldTwentyFiveCellaFiveFivebceTwoSevenNinefZeroFourThreeZeroThreea}{41.3}
\newcommand{\TableOneAccGoldTwentyFiveCellfeebdbafdFiveOneOneaZeroeThree}{24.3}
\newcommand{\TableOneAccGoldTwentyFiveCellTwofTwoabNineOneSevenSevendeFiveThreecOneEight}{40.6}
\newcommand{\TableOneAccGoldTwentyFiveCellOneTwobOnedZerocThreeTwodFouraSixEightNinee}{37.5}
\newcommand{\TableOneAccGoldTwentyFiveCellTwoTwoZerobFiveTwobOneFourZerodSixEightacEight}{40.6}
\newcommand{\TableOneAccGoldTwentyFiveCellTwoEightNineSevenSixfFourTwobNineOneOneeOnebEight}{34.4}
\newcommand{\TableOneAccGoldTwentyFiveCellTwocFourZeroSixbFourEightSixTwoTwoSevenceSevene}{32.8}
\newcommand{\TableOneAccGoldTwentyFiveCellcNinecdOnebEightbEightTwoSixSevenSixeFiveZero}{25.0}
\newcommand{\TableOneAccGoldTwentyFiveCellSevenFivefFivedOneTwoSixSixbThreeaSeveneSevenOne}{28.1}
\newcommand{\TableOneAccGoldTwentyFiveCellSevenFiveSeveneaEightOnecSixbaEightOneFouraOne}{31.2}
\newcommand{\TableOneAccGoldTwentyFiveCellfbThreeEightSixFourThreecSevenNinedfaZerocd}{25.0}
\newcommand{\TableOneAccGoldTwentyFiveCelleThreeNinecfdceSixcFiveeOneZeroOned}{23.4}
\newcommand{\TableOneAccGoldTwentyFiveCellddcTwoEighteOneaebcEightOnebFourb}{18.8}
\newcommand{\TableOneAccGoldTwentyFiveCellbcFourTwoOneOneSevendThreeOnefZeroFivefZeroNine}{25.0}
\newcommand{\TableOneAccGoldTwentyFiveCellTwoffdFiveThreeTwoFourbZeroTwoFourEightcaNine}{17.2}
\newcommand{\TableOneAccGoldTwentyFiveCelladfbFourThreeNineebbfcFourOneeb}{25.0}
\newcommand{\TableOneAccGoldTwentyFiveCellSevenbSevenZeroccbfEightTwoaSixcEightaFour}{21.9}
\newcommand{\TableOneAccGoldTwentyFiveCellZeroSixZerodEightFourSixdThreeEightfbZeroOneNineThree}{21.9}
\newcommand{\TableOneAccGoldTwentyFiveCellEightSevenOneTwoThreeSevenZeroZeroNineEightdcdTwoSixSix}{12.5}
\newcommand{\SupplementConfidenceLevelPercent}{95}
\newcommand{\SupplementBootstrapResamples}{10{,}000}
\newcommand{\SupplementCrossfitFolds}{5}
\newcommand{\SupplementBenchmarkSourceTaskCount}{375}

\newcommand{\SupplementBenchmarkSelectionSystemCount}{8}

\newcommand{\SupplementChallengeCoreTaskCount}{50}
\newcommand{\SupplementChallengeExtensionTaskCount}{14}
\newcommand{\SupplementHoldoutWorldsPerTask}{5}
\newcommand{\SupplementHoldoutSeedOffset}{10{,}000}
\newcommand{\SupplementBenchmarkHoldoutRequestedWorldCount}{1{,}500}
\newcommand{\SupplementBenchmarkHoldoutGeneratedWorldCount}{1{,}465}
\newcommand{\SupplementBenchmarkHoldoutTaskCoverage}{294}
\newcommand{\SupplementChallengeHoldoutRequestedWorldCount}{320}
\newcommand{\SupplementChallengeHoldoutGeneratedWorldCount}{313}
\newcommand{\SupplementChallengeHoldoutTaskCoverage}{63}
\newcommand{\SuppDifficultyBenchmarkBothN}{274}
\newcommand{\SuppDifficultyBenchmarkBothMismatchPercent}{6.1}
\newcommand{\SuppDifficultyBenchmarkBothRepeatSuccessPercent}{91.7}
\newcommand{\SuppDifficultyBenchmarkHFOnlyN}{124}
\newcommand{\SuppDifficultyBenchmarkHFOnlyMismatchPercent}{32.1}
\newcommand{\SuppDifficultyBenchmarkHFOnlyRepeatSuccessPercent}{30.6}
\newcommand{\SuppDifficultyChallengeBothN}{129}
\newcommand{\SuppDifficultyChallengeBothMismatchPercent}{0.0}
\newcommand{\SuppDifficultyChallengeBothRepeatSuccessPercent}{88.1}
\newcommand{\SuppDifficultyChallengeHFOnlyN}{53}
\newcommand{\SuppDifficultyChallengeHFOnlyMismatchPercent}{32.1}
\newcommand{\SuppDifficultyChallengeHFOnlyRepeatSuccessPercent}{32.2}
\newcommand{\SuppRepairRowAPairs}{1{,}036}
\newcommand{\SuppRepairRowAReducedPercent}{96.9}
\newcommand{\SuppRepairRowAExactPercent}{5.4}
\newcommand{\SuppRepairRowBPairs}{77}
\newcommand{\SuppRepairRowBReducedPercent}{83.1}
\newcommand{\SuppRepairRowBExactPercent}{7.8}
\newcommand{\SuppRepairRowCPairs}{98}
\newcommand{\SuppRepairRowCReducedPercent}{88.8}
\newcommand{\SuppRepairRowCExactPercent}{5.1}
\newcommand{\SuppRepairRowDPairs}{62}
\newcommand{\SuppRepairRowDReducedPercent}{98.4}
\newcommand{\SuppRepairRowDExactPercent}{4.8}
\newcommand{\SuppRepairRowEPairs}{173}
\newcommand{\SuppRepairRowEReducedPercent}{91.9}
\newcommand{\SuppRepairRowEExactPercent}{2.9}
\newcommand{\SuppRepairRowFPairs}{194}
\newcommand{\SuppRepairRowFReducedPercent}{89.2}
\newcommand{\SuppRepairRowFExactPercent}{2.1}
\newcommand{\SuppRepairRowGPairs}{256}
\newcommand{\SuppRepairRowGReducedPercent}{98.0}
\newcommand{\SuppRepairRowGExactPercent}{1.2}
\newcommand{\MainControlComparisonTable}{%
\begin{table*}[t]
\centering
\small
\setlength{\tabcolsep}{1.5pt}
\begin{tabular}{llrrrrrrr}
\toprule
Model & Condition & R1 & Valid & \shortstack{Acc@\\gold+25} & H-exact & \shortstack{AST mean\\pre$\rightarrow$post} & \shortstack{Gain vs no-sym.\\(pp), 95\% CI} & \shortstack{LLM\\calls} \\
\midrule
\multicolumn{9}{l}{\textbf{Benchmark300} ($N=300$)}\\
DeepSeek V4 Pro & HF & 17.3\% & \textbf{67.3\%} & \TableOneAccGoldTwentyFiveCelleZeroFourSeveneThreeOneSixThreeadaThreeaEighta{}\% & 35.7\% & 60.5$\rightarrow$45.3 & +16.3 [11.3, 21.7] & 3.20 \\
 & No symbolic & 17.3\% & 51.0\% & \TableOneAccGoldTwentyFiveCellZeroTwoFivedcdFourfaebOneSevenEightae{}\% & 33.3\% & 37.3 & --- & 4.23 \\
Gemini 3.5 Flash & HF & 24.0\% & 65.3\% & \textbf{\TableOneAccGoldTwentyFiveCelldSixTwocSixSixSixTwoSevenFourTwoZeroThreeZeroThreeFour{}\%} & \textbf{45.3\%} & 28.3$\rightarrow$24.8 & +19.3 [14.3, 24.3] & 3.09 \\
 & No symbolic & 24.0\% & 46.0\% & \TableOneAccGoldTwentyFiveCellaEightZerobcEightOnefOneOneNineSevenbThreeThreeNine{}\% & 41.0\% & 16.6 & --- & 4.17 \\
Kimi K2.7 Code & HF & \textbf{27.0\%} & 64.7\% & \TableOneAccGoldTwentyFiveCellThreeSevenFiveFivecFourOneNineZeroFourTwoFourdZeroSixd{}\% & 39.0\% & 48.7$\rightarrow$37.7 & +16.3 [11.7, 21.3] & 3.16 \\
 & No symbolic & \textbf{27.0\%} & 48.3\% & \TableOneAccGoldTwentyFiveCellNineeTwodNineaTwoTwoadaThreeOneSevenSevenSix{}\% & 34.7\% & 33.8 & --- & 4.05 \\
Kimi K2.6 & HF & 15.0\% & 54.0\% & \TableOneAccGoldTwentyFiveCellaFiveFivebceTwoSevenNinefZeroFourThreeZeroThreea{}\% & 33.3\% & 36.4$\rightarrow$29.6 & --- & 3.74 \\
Grok 4.3 & HF & 4.7\% & 29.0\% & \TableOneAccGoldTwentyFiveCellfeebdbafdFiveOneOneaZeroeThree{}\% & 17.3\% & 30.2$\rightarrow$25.0 & --- & 4.41 \\
\midrule
\multicolumn{9}{l}{\textbf{Challenge64} ($N=64$)}\\
GPT-5.6 Sol (xhigh) & HF & \textbf{57.8\%} & \textbf{89.1\%} & \textbf{\TableOneAccGoldTwentyFiveCellTwofTwoabNineOneSevenSevendeFiveThreecOneEight{}\%} & \textbf{32.8\%} & 256.0$\rightarrow$197.3 & +10.9 [3.1, 20.3] & 2.09 \\
 & No symbolic & \textbf{57.8\%} & 78.1\% & \TableOneAccGoldTwentyFiveCellOneTwobOnedZerocThreeTwodFouraSixEightNinee{}\% & 29.7\% & 226.1 & --- & 2.52 \\
GPT-5.6 Terra (xhigh) & HF & 37.5\% & 81.2\% & \textbf{\TableOneAccGoldTwentyFiveCellTwoTwoZerobFiveTwobOneFourZerodSixEightacEight{}\%} & \textbf{32.8\%} & 201.5$\rightarrow$156.8 & +25.0 [14.1, 35.9] & 2.78 \\
 & No symbolic & 37.5\% & 56.2\% & \TableOneAccGoldTwentyFiveCellTwoEightNineSevenSixfFourTwobNineOneOneeOnebEight{}\% & 28.1\% & 159.7 & --- & 3.58 \\
Fable 5 (medium; low) & HF & 40.6\% & 75.0\% & \TableOneAccGoldTwentyFiveCellTwocFourZeroSixbFourEightSixTwoTwoSevenceSevene{}\% & 26.6\% & 108.7$\rightarrow$92.9 & --- & 2.36 \\
Grok 4.5 (high) & HF & 17.2\% & 51.6\% & \TableOneAccGoldTwentyFiveCellcNinecdOnebEightbEightTwoSixSevenSixeFiveZero{}\% & 18.8\% & 141.5$\rightarrow$101.5 & +14.1 [1.6, 26.6] & 4.42 \\
 & No symbolic & 17.2\% & 37.5\% & \TableOneAccGoldTwentyFiveCellSevenFivefFivedOneTwoSixSixbThreeaSeveneSevenOne{}\% & 25.0\% & 35.0 & --- & 4.61 \\
Kimi K3 & HF & 10.9\% & 48.4\% & \TableOneAccGoldTwentyFiveCellSevenFiveSeveneaEightOnecSixbaEightOneFouraOne{}\% & 23.4\% & 85.7$\rightarrow$72.9 & --- & 4.33 \\
GPT-5.6 Luna (xhigh) & HF & 23.4\% & 46.9\% & \TableOneAccGoldTwentyFiveCellfbThreeEightSixFourThreecSevenNinedfaZerocd{}\% & 18.8\% & 175.5$\rightarrow$111.6 & --- & 4.05 \\
DeepSeek V4 Pro & HF & 9.4\% & 34.4\% & \TableOneAccGoldTwentyFiveCelleThreeNinecfdceSixcFiveeOneZeroOned{}\% & 18.8\% & 86.0$\rightarrow$54.8 & +9.4 [-1.6, 20.3] & 4.58 \\
 & No symbolic & 9.4\% & 25.0\% & \TableOneAccGoldTwentyFiveCellddcTwoEighteOneaebcEightOnebFourb{}\% & 15.6\% & 34.8 & --- & 5.16 \\
Kimi K2.7 Code & HF & 7.8\% & 34.4\% & \TableOneAccGoldTwentyFiveCellbcFourTwoOneOneSevendThreeOnefZeroFivefZeroNine{}\% & 18.8\% & 58.9$\rightarrow$49.9 & +15.6 [6.2, 25.0] & 4.70 \\
 & No symbolic & 7.8\% & 18.8\% & \TableOneAccGoldTwentyFiveCellTwoffdFiveThreeTwoFourbZeroTwoFourEightcaNine{}\% & 15.6\% & 24.8 & --- & 5.28 \\
Gemini 3.5 Flash & HF & 10.9\% & 28.1\% & \TableOneAccGoldTwentyFiveCelladfbFourThreeNineebbfcFourOneeb{}\% & 20.3\% & 24.9$\rightarrow$24.2 & +6.2 [-3.1, 15.6] & 4.58 \\
 & No symbolic & 10.9\% & 21.9\% & \TableOneAccGoldTwentyFiveCellSevenbSevenZeroccbfEightTwoaSixcEightaFour{}\% & 18.8\% & 16.2 & --- & 5.20 \\
Kimi K2.6 & HF & 4.7\% & 23.4\% & \TableOneAccGoldTwentyFiveCellZeroSixZerodEightFourSixdThreeEightfbZeroOneNineThree{}\% & 17.2\% & 25.4$\rightarrow$24.1 & --- & 5.08 \\
Grok 4.3 & HF & 3.1\% & 12.5\% & \TableOneAccGoldTwentyFiveCellEightSevenOneTwoThreeSevenZeroZeroNineEightdcdTwoSixSix{}\% & 9.4\% & 18.9$\rightarrow$18.6 & --- & 5.36 \\
\bottomrule
\end{tabular}
\caption{Exact validity and formula quality under Hypothesis Frontier (HF) and repeated generation. R1 is direct Round~1 validity. Valid, Acc@gold+25, and H-exact use the full task denominator. Acc requires a train-valid final formula no more than 25 AST nodes above the planted reference, used only for reporting; H-exact requires agreement on every holdout world. AST pre$\rightarrow$post shows final exact simplification of train-valid formulas; no-symbolic rows show direct AST. Gain and paired 95\% confidence intervals compare matched conditions. Calls are mean reported LLM calls per task, excluding API retries. Fable 5 uses medium effort with a low-effort fallback.}
\label{tab:main-control-comparison}
\end{table*}
}

\newcommand{\SymbolicOnlyBaselineTable}{%
\begin{table*}[t]
\centering
\small
\begin{minipage}[t]{0.545\textwidth}
\centering
\textbf{(a) Symbolic-only validity and quality}\par
\smallskip
\begin{tabular}{@{}llrrrr@{}}
\toprule
View & Method & Valid & Acc@gold+25 & H-exact & AST med. \\
\midrule
B300 & \texttt{prenex} &
\ZThreePrenexBenchmarkValidPercent{}\% &
\ZThreePrenexBenchmarkAccTwentyFivePercent{}\% &
\ZThreePrenexBenchmarkHExactPercent{}\% &
\ZThreePrenexBenchmarkASTMedian{} \\
B300 & \texttt{ad-mix} &
\ZThreeAdMixBenchmarkValidPercent{}\% &
\ZThreeAdMixBenchmarkAccTwentyFivePercent{}\% &
\ZThreeAdMixBenchmarkHExactPercent{}\% &
\ZThreeAdMixBenchmarkASTMedian{} \\
C64 & \texttt{prenex} &
\ZThreePrenexChallengeValidPercent{}\% &
\ZThreePrenexChallengeAccTwentyFivePercent{}\% &
\ZThreePrenexChallengeHExactPercent{}\% &
\ZThreePrenexChallengeASTMedian{} \\
C64 & \texttt{ad-mix} &
\ZThreeAdMixChallengeValidPercent{}\% &
\ZThreeAdMixChallengeAccTwentyFivePercent{}\% &
\ZThreeAdMixChallengeHExactPercent{}\% &
\ZThreeAdMixChallengeASTMedian{} \\
\bottomrule
\end{tabular}
\end{minipage}%
\begin{minipage}[t]{0.455\textwidth}
\centering
\textbf{(b) Coverage overlap with Hypothesis Frontier}\par
\smallskip
\begin{tabular}{@{}llrrrr@{}}
\toprule
View & Z3 & Both & HF only & Z3 only & Union \\
\midrule
B300 & \texttt{prenex} &
\ZThreePrenexBenchmarkBothCount{} &
\ZThreePrenexBenchmarkHFOnlyCount{} &
\ZThreePrenexBenchmarkZThreeOnlyCount{} &
\ZThreePrenexBenchmarkUnionCount{} \\
B300 & \texttt{ad-mix} &
\ZThreeAdMixBenchmarkBothCount{} &
\ZThreeAdMixBenchmarkHFOnlyCount{} &
\ZThreeAdMixBenchmarkZThreeOnlyCount{} &
\ZThreeAdMixBenchmarkUnionCount{} \\
C64 & \texttt{prenex} &
\ZThreePrenexChallengeBothCount{} &
\ZThreePrenexChallengeHFOnlyCount{} &
\ZThreePrenexChallengeZThreeOnlyCount{} &
\ZThreePrenexChallengeUnionCount{} \\
C64 & \texttt{ad-mix} &
\ZThreeAdMixChallengeBothCount{} &
\ZThreeAdMixChallengeHFOnlyCount{} &
\ZThreeAdMixChallengeZThreeOnlyCount{} &
\ZThreeAdMixChallengeUnionCount{} \\
\bottomrule
\end{tabular}
\end{minipage}
\par\medskip
\begin{minipage}[t]{\textwidth}
\centering
\textbf{(c) Prenex-first Hypothesis Frontier by configuration}\par
\smallskip
\begin{tabular}{@{}llrrrr@{}}
\toprule
View & Model & Valid & H-exact & \shortstack{LLM calls\\per task} & \shortstack{Calls saved\\(\%)} \\
\midrule
B300 & DeepSeek V4 Pro & \CascadePrenexHFBThreeZeroZeroDeepseekOnePromptVsRepeatValidPercent{}\% & \CascadePrenexHFBThreeZeroZeroDeepseekOnePromptVsRepeatHExactPercent{}\% & \CascadePrenexHFBThreeZeroZeroDeepseekOnePromptVsRepeatCallsPerTask{} & \CascadePrenexHFBThreeZeroZeroDeepseekOnePromptVsRepeatCallsSavedPercent{}\% \\
B300 & Gemini 3.5 Flash & \CascadePrenexHFBThreeZeroZeroGeminiOnePromptVsRepeatValidPercent{}\% & \CascadePrenexHFBThreeZeroZeroGeminiOnePromptVsRepeatHExactPercent{}\% & \CascadePrenexHFBThreeZeroZeroGeminiOnePromptVsRepeatCallsPerTask{} & \CascadePrenexHFBThreeZeroZeroGeminiOnePromptVsRepeatCallsSavedPercent{}\% \\
B300 & Kimi K2.7 Code & \CascadePrenexHFBThreeZeroZeroKimiTwoSevenValidPercent{}\% & \CascadePrenexHFBThreeZeroZeroKimiTwoSevenHExactPercent{}\% & \CascadePrenexHFBThreeZeroZeroKimiTwoSevenCallsPerTask{} & \CascadePrenexHFBThreeZeroZeroKimiTwoSevenCallsSavedPercent{}\% \\
\addlinespace[2pt]
C64 & GPT-5.6 Sol (xhigh) & \CascadePrenexHFCSixFourNativeSolOnePromptVsRepeatValidPercent{}\% & \CascadePrenexHFCSixFourNativeSolOnePromptVsRepeatHExactPercent{}\% & \CascadePrenexHFCSixFourNativeSolOnePromptVsRepeatCallsPerTask{} & \CascadePrenexHFCSixFourNativeSolOnePromptVsRepeatCallsSavedPercent{}\% \\
C64 & GPT-5.6 Terra (xhigh) & \CascadePrenexHFCSixFourNativeTerraOnePromptVsRepeatValidPercent{}\% & \CascadePrenexHFCSixFourNativeTerraOnePromptVsRepeatHExactPercent{}\% & \CascadePrenexHFCSixFourNativeTerraOnePromptVsRepeatCallsPerTask{} & \CascadePrenexHFCSixFourNativeTerraOnePromptVsRepeatCallsSavedPercent{}\% \\
C64 & Fable 5 (medium; low) & \CascadePrenexHFCSixFourFableFiveValidPercent{}\% & \CascadePrenexHFCSixFourFableFiveHExactPercent{}\% & \CascadePrenexHFCSixFourFableFiveCallsPerTask{} & \CascadePrenexHFCSixFourFableFiveCallsSavedPercent{}\% \\
C64 & Grok 4.5 (high) & \CascadePrenexHFCSixFourNativeGrokFourFiveOnePromptVsRepeatValidPercent{}\% & \CascadePrenexHFCSixFourNativeGrokFourFiveOnePromptVsRepeatHExactPercent{}\% & \CascadePrenexHFCSixFourNativeGrokFourFiveOnePromptVsRepeatCallsPerTask{} & \CascadePrenexHFCSixFourNativeGrokFourFiveOnePromptVsRepeatCallsSavedPercent{}\% \\
C64 & GPT-5.6 Luna (xhigh) & \CascadePrenexHFCSixFourLunaValidPercent{}\% & \CascadePrenexHFCSixFourLunaHExactPercent{}\% & \CascadePrenexHFCSixFourLunaCallsPerTask{} & \CascadePrenexHFCSixFourLunaCallsSavedPercent{}\% \\
\bottomrule
\end{tabular}
\end{minipage}
\caption{Independent symbolic synthesis, overlap with Hypothesis
Frontier (HF), and a prenex-first cascade. Panel (a) evaluates
selected Z3 formulas as returned; metrics use the full denominator
except AST median, which is over train-valid formulas. Panel (b) gives
task--model overlap. Panel (c) reports the complete prenex-first
cascade configurations; accepted Z3 formulas remain unchanged and
routed HF solutions use their final simplified formulas. Calls count
at most one call per task and round and exclude API retries;
calls saved also excludes symbolic-front-end computation. Appendix
Table~\ref{tab:supp-symbolic-first} gives the complete cascade
comparison.}
\label{tab:symbolic-only-baselines}
\end{table*}
}

\renewcommand{\MainControlComparisonTable}{%
\begin{table*}[!t]
\centering
\small
\setlength{\tabcolsep}{1.5pt}
\begin{tabular}{llrrrrrrr}
\toprule
Model & Condition & R1 & Valid & \shortstack{Acc@\\gold+25} & H-exact & \shortstack{AST mean\\pre$\rightarrow$post} & \shortstack{Gain vs no-sym.\\(pp), 95\% CI} & \shortstack{LLM\\calls} \\
\midrule
\multicolumn{9}{l}{\textbf{Benchmark300} ($N=300$)}\\
DeepSeek V4 Pro & HF & 17.3\% & \textbf{67.3\%} & \TableOneAccGoldTwentyFiveCelleZeroFourSeveneThreeOneSixThreeadaThreeaEighta{}\% & 35.7\% & 60.5$\rightarrow$45.3 & +16.3 [11.3, 21.7] & 3.20 \\
 & No symbolic & 17.3\% & 51.0\% & \TableOneAccGoldTwentyFiveCellZeroTwoFivedcdFourfaebOneSevenEightae{}\% & 33.3\% & 37.3 & --- & 4.23 \\
Gemini 3.5 Flash & HF & 24.0\% & 65.3\% & \textbf{\TableOneAccGoldTwentyFiveCelldSixTwocSixSixSixTwoSevenFourTwoZeroThreeZeroThreeFour{}\%} & \textbf{45.3\%} & 28.3$\rightarrow$24.8 & +19.3 [14.3, 24.3] & 3.09 \\
 & No symbolic & 24.0\% & 46.0\% & \TableOneAccGoldTwentyFiveCellaEightZerobcEightOnefOneOneNineSevenbThreeThreeNine{}\% & 41.0\% & 16.6 & --- & 4.17 \\
Kimi K2.7 Code & HF & \textbf{27.0\%} & 64.7\% & \TableOneAccGoldTwentyFiveCellThreeSevenFiveFivecFourOneNineZeroFourTwoFourdZeroSixd{}\% & 39.0\% & 48.7$\rightarrow$37.7 & +16.3 [11.7, 21.3] & 3.16 \\
 & No symbolic & \textbf{27.0\%} & 48.3\% & \TableOneAccGoldTwentyFiveCellNineeTwodNineaTwoTwoadaThreeOneSevenSevenSix{}\% & 34.7\% & 33.8 & --- & 4.05 \\
Kimi K2.6 & HF & 15.0\% & 54.0\% & \TableOneAccGoldTwentyFiveCellaFiveFivebceTwoSevenNinefZeroFourThreeZeroThreea{}\% & 33.3\% & 36.4$\rightarrow$29.6 & --- & 3.74 \\
Grok 4.3 & HF & 4.7\% & 29.0\% & \TableOneAccGoldTwentyFiveCellfeebdbafdFiveOneOneaZeroeThree{}\% & 17.3\% & 30.2$\rightarrow$25.0 & --- & 4.41 \\
\midrule
\multicolumn{9}{l}{\textbf{Challenge64} ($N=64$)}\\
GPT-5.6 Sol (xhigh) & HF & \textbf{57.8\%} & \textbf{89.1\%} & \textbf{\TableOneAccGoldTwentyFiveCellTwofTwoabNineOneSevenSevendeFiveThreecOneEight{}\%} & \textbf{32.8\%} & 256.0$\rightarrow$197.3 & +10.9 [3.1, 20.3] & 2.09 \\
 & No symbolic & \textbf{57.8\%} & 78.1\% & \TableOneAccGoldTwentyFiveCellOneTwobOnedZerocThreeTwodFouraSixEightNinee{}\% & 29.7\% & 226.1 & --- & 2.52 \\
GPT-5.6 Terra (xhigh) & HF & 37.5\% & 81.2\% & \textbf{\TableOneAccGoldTwentyFiveCellTwoTwoZerobFiveTwobOneFourZerodSixEightacEight{}\%} & \textbf{32.8\%} & 201.5$\rightarrow$156.8 & +25.0 [14.1, 35.9] & 2.78 \\
 & No symbolic & 37.5\% & 56.2\% & \TableOneAccGoldTwentyFiveCellTwoEightNineSevenSixfFourTwobNineOneOneeOnebEight{}\% & 28.1\% & 159.7 & --- & 3.58 \\
Fable 5 (medium; low) & HF & 40.6\% & 75.0\% & \TableOneAccGoldTwentyFiveCellTwocFourZeroSixbFourEightSixTwoTwoSevenceSevene{}\% & 26.6\% & 108.7$\rightarrow$92.9 & --- & 2.36 \\
Grok 4.5 (high) & HF & 17.2\% & 51.6\% & \TableOneAccGoldTwentyFiveCellcNinecdOnebEightbEightTwoSixSevenSixeFiveZero{}\% & 18.8\% & 141.5$\rightarrow$101.5 & +14.1 [1.6, 26.6] & 4.42 \\
 & No symbolic & 17.2\% & 37.5\% & \TableOneAccGoldTwentyFiveCellSevenFivefFivedOneTwoSixSixbThreeaSeveneSevenOne{}\% & 25.0\% & 35.0 & --- & 4.61 \\
Kimi K3 & HF & 10.9\% & 48.4\% & \TableOneAccGoldTwentyFiveCellSevenFiveSeveneaEightOnecSixbaEightOneFouraOne{}\% & 23.4\% & 85.7$\rightarrow$72.9 & --- & 4.33 \\
GPT-5.6 Luna (xhigh) & HF & 23.4\% & 46.9\% & \TableOneAccGoldTwentyFiveCellfbThreeEightSixFourThreecSevenNinedfaZerocd{}\% & 18.8\% & 175.5$\rightarrow$111.6 & --- & 4.05 \\
DeepSeek V4 Pro & HF & 9.4\% & 34.4\% & \TableOneAccGoldTwentyFiveCelleThreeNinecfdceSixcFiveeOneZeroOned{}\% & 18.8\% & 86.0$\rightarrow$54.8 & +9.4 [-1.6, 20.3] & 4.58 \\
 & No symbolic & 9.4\% & 25.0\% & \TableOneAccGoldTwentyFiveCellddcTwoEighteOneaebcEightOnebFourb{}\% & 15.6\% & 34.8 & --- & 5.16 \\
Kimi K2.7 Code & HF & 7.8\% & 34.4\% & \TableOneAccGoldTwentyFiveCellbcFourTwoOneOneSevendThreeOnefZeroFivefZeroNine{}\% & 18.8\% & 58.9$\rightarrow$49.9 & +15.6 [6.2, 25.0] & 4.70 \\
 & No symbolic & 7.8\% & 18.8\% & \TableOneAccGoldTwentyFiveCellTwoffdFiveThreeTwoFourbZeroTwoFourEightcaNine{}\% & 15.6\% & 24.8 & --- & 5.28 \\
Gemini 3.5 Flash & HF & 10.9\% & 28.1\% & \TableOneAccGoldTwentyFiveCelladfbFourThreeNineebbfcFourOneeb{}\% & 20.3\% & 24.9$\rightarrow$24.2 & +6.2 [-3.1, 15.6] & 4.58 \\
 & No symbolic & 10.9\% & 21.9\% & \TableOneAccGoldTwentyFiveCellSevenbSevenZeroccbfEightTwoaSixcEightaFour{}\% & 18.8\% & 16.2 & --- & 5.20 \\
Kimi K2.6 & HF & 4.7\% & 23.4\% & \TableOneAccGoldTwentyFiveCellZeroSixZerodEightFourSixdThreeEightfbZeroOneNineThree{}\% & 17.2\% & 25.4$\rightarrow$24.1 & --- & 5.08 \\
Grok 4.3 & HF & 3.1\% & 12.5\% & \TableOneAccGoldTwentyFiveCellEightSevenOneTwoThreeSevenZeroZeroNineEightdcdTwoSixSix{}\% & 9.4\% & 18.9$\rightarrow$18.6 & --- & 5.36 \\
\bottomrule
\end{tabular}
\caption{Exact validity and formula quality under Hypothesis Frontier (HF) and repeated generation. R1 is direct Round~1 validity. Valid, Acc@gold+25, and H-exact use the full task denominator. Acc requires a train-valid final formula no more than 25 AST nodes above the planted reference, used only for reporting; H-exact requires agreement on every holdout world. AST pre$\rightarrow$post shows final exact simplification of train-valid formulas; no-symbolic rows show direct AST. Gain and paired 95\% confidence intervals compare matched conditions. Calls are mean reported LLM calls per task, excluding API retries. Fable 5 uses medium effort with a low-effort fallback.}
\label{tab:main-control-comparison}
\end{table*}
}
\newcommand{\SupplementAnalysisConfigurationTable}{%
\begin{table*}[t]
\centering
\small
\setlength{\tabcolsep}{5pt}
\begin{tabular}{@{}p{0.22\textwidth}p{0.34\textwidth}p{0.38\textwidth}@{}}
\toprule
Analysis & Benchmark300 & Challenge64 \\
\midrule
Main comparison and Figure 1 & DeepSeek V4 Pro, Gemini 3.5 Flash, Kimi K2.7 Code & DeepSeek V4 Pro, Gemini 3.5 Flash, GPT-5.6 Sol (xhigh), GPT-5.6 Terra (xhigh), Grok 4.5 (high), Kimi K2.7 Code \\
Parent-linked repair analyses & Grok 4.3 & Fable 5 (medium; low), GPT-5.6 Luna (xhigh), GPT-5.6 Sol (xhigh), GPT-5.6 Terra (xhigh), Grok 4.3, Grok 4.5 (high) \\
Paired quality, difficulty, and Z3 overlap & DeepSeek V4 Pro, Gemini 3.5 Flash & DeepSeek V4 Pro, Gemini 3.5 Flash, GPT-5.6 Sol (xhigh), GPT-5.6 Terra (xhigh), Grok 4.5 (high) \\
Main prenex-first panel & DeepSeek V4 Pro, Gemini 3.5 Flash, Kimi K2.7 Code & Fable 5 (medium; low), GPT-5.6 Luna (xhigh), GPT-5.6 Sol (xhigh), GPT-5.6 Terra (xhigh), Grok 4.5 (high) \\
Symbolic-first comparison & DeepSeek V4 Pro, Gemini 3.5 Flash & DeepSeek V4 Pro, Gemini 3.5 Flash, GPT-5.6 Sol (xhigh), GPT-5.6 Terra (xhigh), Grok 4.5 (high) \\
Final exact simplification & All reported LLM--symbolic configurations & All reported LLM--symbolic configurations \\
\bottomrule
\end{tabular}
\caption{Configurations included in each analysis. Each row contains every configuration that supplies the information required by that analysis. HF denotes Hypothesis Frontier. Figure~1 uses the complete matched-control set, including its first-source decomposition; the separate repair analyses require a recorded parent for each symbolic edit. Paired analyses require Hypothesis Frontier and repeated-generation outcomes for the same model and tasks. Final exact simplification applies only to train-valid formulas from LLM--symbolic conditions; no-symbolic formulas remain unchanged.}
\label{tab:supp-analysis-configurations}
\end{table*}
}

\newcommand{\SupplementRoundAndSimplificationTable}{%
\begin{table*}[t]
\centering
\small
\setlength{\tabcolsep}{5pt}
\begin{tabular}{@{}llrrrrrr@{}}
\toprule
View & Selection & R1 & R2 & R3 & R4 & R5 & R6 \\
\midrule
B300 & HF & 22.8/35.3 & 53.4/55.9 & 59.0/59.9 & 61.1/62.6 & 63.4/64.1 & 65.2/65.8 \\
B300 & Repeated & 22.8/-- & 32.8/-- & 39.1/-- & 43.3/-- & 46.9/-- & 48.4/-- \\
\addlinespace[2pt]
C64 & HF & 23.4/27.3 & 37.8/39.1 & 43.8/44.8 & 46.1/46.4 & 49.5/50.0 & 52.6/53.1 \\
C64 & Repeated & 23.4/-- & 29.9/-- & 33.6/-- & 35.4/-- & 38.5/-- & 39.6/-- \\
\bottomrule
\end{tabular}
\caption{Cumulative train validity by round, pooled over the matched configurations within each benchmark. Each cell is pre/post in percent. For HF, pre combines the incoming verified frontier with newly valid LLM proposals from the current round; post additionally includes symbolic descendants accepted in that round. Repeated generation reports cumulative direct validity and has no symbolic stage, so its post entry is shown as --.}
\label{tab:supp-rounds-simplification}
\end{table*}
}

\newcommand{\SupplementBenchmarkConfigurationTrajectoriesTable}{%
\begin{table*}[t]
\centering
\small
\setlength{\tabcolsep}{4.5pt}
\begin{tabular}{@{}llrrrrrr@{}}
\toprule
Model & Condition & R1 & R2 & R3 & R4 & R5 & R6 \\
\midrule
DeepSeek V4 Pro & HF & 17.3/28.7 & 55.0/55.3 & 60.7/62.0 & 63.7/64.0 & 64.7/65.7 & 67.3/67.3 \\
 & 3-formula & 17.3/28.7 & 38.0/42.0 & 47.0/49.3 & 53.3/54.7 & 55.3/56.3 & 56.3/56.3 \\
 & Repeated & 17.3/-- & 31.0/-- & 37.3/-- & 42.3/-- & 48.7/-- & 51.0/-- \\
\addlinespace[1.5pt]
Gemini 3.5 Flash & HF & 24.0/34.7 & 58.0/60.3 & 61.7/61.7 & 63.0/63.3 & 64.0/64.0 & 65.3/65.3 \\
 & 3-formula & 24.0/34.7 & 38.7/40.7 & 45.7/48.3 & 52.3/54.0 & 54.7/55.0 & 55.7/56.7 \\
 & Repeated & 24.0/-- & 33.0/-- & 39.0/-- & 42.3/-- & 45.0/-- & 46.0/-- \\
\addlinespace[1.5pt]
Kimi K2.7 Code & HF & 27.0/42.7 & 47.3/52.0 & 54.7/56.0 & 56.7/60.3 & 61.7/62.7 & 63.0/64.7 \\
 & Repeated & 27.0/-- & 34.3/-- & 41.0/-- & 45.3/-- & 47.0/-- & 48.3/-- \\
\addlinespace[1.5pt]
Kimi K2.6 & HF & 15.0/29.0 & 41.0/43.7 & 46.0/47.3 & 48.7/50.3 & 51.7/53.0 & 53.3/54.0 \\
 & 3-formula & 15.0/29.0 & 32.3/36.7 & 38.0/41.7 & 44.0/45.7 & 46.7/48.0 & 48.3/50.0 \\
\addlinespace[1.5pt]
Grok 4.3 & HF & 4.7/16.3 & 18.3/19.3 & 21.7/24.0 & 24.7/25.0 & 25.3/27.7 & 28.0/29.0 \\
\bottomrule
\end{tabular}
\caption{Configuration-level cumulative train validity on Benchmark300 for the HF, three-formula, and repeated-generation configurations available for this view. Each cell is pre/post in percent. For HF and three-formula configurations, pre combines the incoming verified frontier with the current round's direct proposals; post adds the symbolic descendants accepted in that round. Repeated generation reports cumulative direct validity and has no post-symbolic stage.}
\label{tab:supp-benchmark-config-trajectories}
\end{table*}
}

\newcommand{\SupplementChallengeConfigurationTrajectoriesTable}{%
\begin{table*}[t]
\centering
\small
\setlength{\tabcolsep}{4.5pt}
\begin{tabular}{@{}llrrrrrr@{}}
\toprule
Model & Condition & R1 & R2 & R3 & R4 & R5 & R6 \\
\midrule
GPT-5.6 Sol (xhigh) & HF & 57.8/57.8 & 73.4/76.6 & 81.2/82.8 & 84.4/84.4 & 89.1/89.1 & 89.1/89.1 \\
 & Repeated & 57.8/-- & 67.2/-- & 73.4/-- & 75.0/-- & 75.0/-- & 78.1/-- \\
\addlinespace[1.5pt]
GPT-5.6 Terra (xhigh) & HF & 37.5/42.2 & 56.2/57.8 & 67.2/68.8 & 68.8/68.8 & 76.6/76.6 & 81.2/81.2 \\
 & Repeated & 37.5/-- & 45.3/-- & 50.0/-- & 53.1/-- & 56.2/-- & 56.2/-- \\
\addlinespace[1.5pt]
Fable 5 (medium; low) & HF & 40.6/45.3 & 60.9/62.5 & 68.8/70.3 & 71.9/71.9 & 73.4/75.0 & 75.0/75.0 \\
\addlinespace[1.5pt]
Grok 4.5 (high) & HF & 17.2/18.8 & 25.0/25.0 & 31.2/34.4 & 37.5/37.5 & 42.2/42.2 & 50.0/51.6 \\
 & Repeated & 17.2/-- & 26.6/-- & 29.7/-- & 29.7/-- & 35.9/-- & 37.5/-- \\
\addlinespace[1.5pt]
Kimi K3 & HF & 10.9/17.2 & 25.0/26.6 & 32.8/34.4 & 43.8/43.8 & 45.3/45.3 & 48.4/48.4 \\
\addlinespace[1.5pt]
GPT-5.6 Luna (xhigh) & HF & 23.4/25.0 & 26.6/29.7 & 35.9/35.9 & 37.5/40.6 & 40.6/40.6 & 46.9/46.9 \\
\addlinespace[1.5pt]
DeepSeek V4 Pro & HF & 9.4/14.1 & 28.1/28.1 & 29.7/29.7 & 29.7/29.7 & 29.7/32.8 & 34.4/34.4 \\
 & 3-formula & 9.4/14.1 & 17.2/20.3 & 21.9/21.9 & 23.4/23.4 & 23.4/23.4 & 23.4/23.4 \\
 & Repeated & 9.4/-- & 15.6/-- & 17.2/-- & 18.8/-- & 23.4/-- & 25.0/-- \\
\addlinespace[1.5pt]
Kimi K2.7 Code & HF & 7.8/17.2 & 20.3/23.4 & 28.1/28.1 & 28.1/29.7 & 31.2/31.2 & 32.8/34.4 \\
 & Repeated & 7.8/-- & 12.5/-- & 14.1/-- & 18.8/-- & 18.8/-- & 18.8/-- \\
\addlinespace[1.5pt]
Gemini 3.5 Flash & HF & 10.9/14.1 & 23.4/23.4 & 25.0/25.0 & 28.1/28.1 & 28.1/28.1 & 28.1/28.1 \\
 & 3-formula & 10.9/14.1 & 17.2/17.2 & 17.2/18.8 & 21.9/21.9 & 21.9/21.9 & 21.9/21.9 \\
 & Repeated & 10.9/-- & 12.5/-- & 17.2/-- & 17.2/-- & 21.9/-- & 21.9/-- \\
\addlinespace[1.5pt]
Kimi K2.6 & HF & 4.7/14.1 & 15.6/15.6 & 18.8/20.3 & 21.9/21.9 & 23.4/23.4 & 23.4/23.4 \\
 & 3-formula & 4.7/14.1 & 15.6/15.6 & 15.6/15.6 & 17.2/21.9 & 21.9/21.9 & 21.9/21.9 \\
\addlinespace[1.5pt]
Grok 4.3 & HF & 3.1/7.8 & 10.9/10.9 & 12.5/12.5 & 12.5/12.5 & 12.5/12.5 & 12.5/12.5 \\
\bottomrule
\end{tabular}
\caption{Configuration-level cumulative train validity on Challenge64 for the HF, three-formula, and repeated-generation configurations available for this view. Each cell is pre/post in percent. For HF and three-formula configurations, pre combines the incoming verified frontier with the current round's direct proposals; post adds the symbolic descendants accepted in that round. Repeated generation reports cumulative direct validity and has no post-symbolic stage.}
\label{tab:supp-challenge-config-trajectories}
\end{table*}
}

\newcommand{\SupplementFormulaQualityTable}{%
\begin{table*}[t]
\centering
\small
\setlength{\tabcolsep}{3pt}
\begin{tabular}{lrrrrr}
\toprule
\multicolumn{6}{l}{\textbf{(a) Fixed-denominator formula quality}}\\
View & Valid & Acc@gold+10 & Acc@gold+25 & Bloat & H-exact \\
\midrule
Benchmark300 & 29.0\% & 20.3\% & 24.3\% & 16.1\% & 17.3\% \\
Challenge64 & 59.4\% & 25.0\% & 29.4\% & 50.4\% & 23.2\% \\
\bottomrule
\end{tabular}
\medskip
\begin{tabular}{lrrrrr}
\toprule
\multicolumn{6}{l}{\textbf{(b) Final exact simplification of train-valid formulas}}\\
View & Smaller / inputs & \shortstack{AST mean\\pre$\rightarrow$post} & Med. removed & \shortstack{H-exact\\pre$\rightarrow$post} & \shortstack{World-exact\\+/=/-} \\
\midrule
Benchmark300 & 461 / 1330 (34.7\%) & 40.9$\rightarrow$32.5 & 12.0 & 30.4\%$\rightarrow$33.6\% & 114/1155/29 \\
Challenge64 & 173 / 379 (45.6\%) & 128.3$\rightarrow$98.4 & 27.0 & 18.9\%$\rightarrow$20.0\% & 20/337/15 \\
\bottomrule
\end{tabular}
\medskip
\begin{tabular}{lrrrrrrr}
\toprule
\multicolumn{8}{l}{\textbf{(c) Holdout task exactness by final formula size}}\\
View & $\leq 15$ & 16--25 & 26--50 & $>50$ & $\leq$gold & gold+1--25 & $>$gold+25 \\
\midrule
Benchmark300 & 100.0\% (40) & 75.0\% (16) & 0.0\% (19) & 0.0\% (9) & 100.0\% (52) & 0.0\% (18) & 0.0\% (14) \\
Challenge64 & 100.0\% (29) & 90.3\% (62) & 8.3\% (24) & 1.9\% (108) & 100.0\% (85) & 8.7\% (23) & 1.7\% (115) \\
\bottomrule
\end{tabular}
\medskip
\begin{tabular}{lrrr}
\toprule
\multicolumn{4}{l}{\textbf{(d) Paired world-exactness advantage of the shorter formula}}\\
View & \shortstack{All pairs\\(pp)} & \shortstack{Excl. shorter = gold\\(pp)} & \shortstack{Both $>$ gold\\(pp)} \\
\midrule
Benchmark300 & \CompactAllDelta_B{} [\CompactAllLo_B{}, \CompactAllHi_B{}] (\CompactAllN_B{}) & \CompactNoGoldDelta_B{} [\CompactNoGoldLo_B{}, \CompactNoGoldHi_B{}] (\CompactNoGoldN_B{}) & \CompactAboveGoldDelta_B{} [\CompactAboveGoldLo_B{}, \CompactAboveGoldHi_B{}] (\CompactAboveGoldN_B{}) \\
Challenge64 & \CompactAllDelta_C{} [\CompactAllLo_C{}, \CompactAllHi_C{}] (\CompactAllN_C{}) & \CompactNoGoldDelta_C{} [\CompactNoGoldLo_C{}, \CompactNoGoldHi_C{}] (\CompactNoGoldN_C{}) & \CompactAboveGoldDelta_C{} [\CompactAboveGoldLo_C{}, \CompactAboveGoldHi_C{}] (\CompactAboveGoldN_C{}) \\
\bottomrule
\end{tabular}
\caption{Formula quality, exact simplification, and holdout exactness. Panels (a) and (c) use the seven primary single-best configurations with complete parent links and no independent symbolic replacement. In panel (a), bloat is conditional on train validity, while the other rates use the full denominator. Panel (b) compares paired train-valid formulas immediately before and after the final exact simplification pass across all symbolic conditions for which both formulas are available. H-exact uses all task--model pairs as the denominator; world-exact improve/same/worsen counts compare the same generated holdout worlds and omit pairs without complete holdout results. Panel (c) relates final formula size to holdout task exactness. Panel (d) compares shorter and longer train-valid formulas for the same task and model across all available conditions; the sensitivity columns remove exact planted-reference matches or retain only pairs in which both formulas exceed the reference size.}
\label{tab:supp-quality-generalization}
\end{table*}
}

\newcommand{\SupplementBenchmarkConfigurationSimplificationTable}{%
\begin{table*}[t]
\centering
\small
\setlength{\tabcolsep}{3.2pt}
\begin{tabular}{@{}llrrrrr@{}}
\toprule
Model & Condition & Smaller/inputs & \shortstack{AST mean\\pre$\rightarrow$post} & \shortstack{Median nodes\\removed} & \shortstack{H-exact\\pre$\rightarrow$post} & \shortstack{World exact\\$+$/=/$-$} \\
\midrule
DeepSeek V4 Pro & HF & 85/202 (42.1\%) & 60.5$\rightarrow$45.3 & 14.0 & 31.0\%$\rightarrow$35.7\% & 19/174/5 \\
 & 3-formula & 70/169 (41.4\%) & 45.3$\rightarrow$35.0 & 13.5 & 27.7\%$\rightarrow$31.3\% & 18/141/6 \\
\addlinespace[1.5pt]
Gemini 3.5 Flash & HF & 43/196 (21.9\%) & 28.3$\rightarrow$24.8 & 12.0 & 42.3\%$\rightarrow$45.3\% & 13/176/2 \\
 & 3-formula & 49/170 (28.8\%) & 30.1$\rightarrow$25.2 & 9.0 & 35.3\%$\rightarrow$38.3\% & 12/154/0 \\
\addlinespace[1.5pt]
Kimi K2.7 Code & HF & 72/194 (37.1\%) & 48.7$\rightarrow$37.7 & 14.0 & 35.7\%$\rightarrow$39.0\% & 14/165/10 \\
\addlinespace[1.5pt]
Kimi K2.6 & HF & 51/162 (31.5\%) & 36.4$\rightarrow$29.6 & 11.0 & 29.7\%$\rightarrow$33.3\% & 17/139/2 \\
 & 3-formula & 56/150 (37.3\%) & 39.0$\rightarrow$31.5 & 13.0 & 25.0\%$\rightarrow$28.7\% & 15/130/2 \\
\addlinespace[1.5pt]
Grok 4.3 & HF & 35/87 (40.2\%) & 30.2$\rightarrow$25.0 & 6.0 & 16.3\%$\rightarrow$17.3\% & 6/76/2 \\
\bottomrule
\end{tabular}
\caption{Per-configuration exact simplification on Benchmark300, expanding panel (b) of Table~\ref{tab:supp-quality-generalization}. Inputs are final train-valid formulas from symbolic conditions; Smaller counts strict AST reductions that preserve every training prediction. AST means are over those inputs. H-exact uses the full task denominator, and world-exact counts compare the same generated holdout worlds before and after simplification.}
\label{tab:supp-benchmark-config-simplification}
\end{table*}
}

\newcommand{\SupplementChallengeConfigurationSimplificationTable}{%
\begin{table*}[t]
\centering
\small
\setlength{\tabcolsep}{3.2pt}
\begin{tabular}{@{}llrrrrr@{}}
\toprule
Model & Condition & Smaller/inputs & \shortstack{AST mean\\pre$\rightarrow$post} & \shortstack{Median nodes\\removed} & \shortstack{H-exact\\pre$\rightarrow$post} & \shortstack{World exact\\$+$/=/$-$} \\
\midrule
GPT-5.6 Sol (xhigh) & HF & 33/57 (57.9\%) & 256.0$\rightarrow$197.3 & 47.0 & 31.2\%$\rightarrow$32.8\% & 3/50/3 \\
\addlinespace[1.5pt]
GPT-5.6 Terra (xhigh) & HF & 26/52 (50.0\%) & 201.5$\rightarrow$156.8 & 48.5 & 32.8\%$\rightarrow$32.8\% & 1/49/1 \\
\addlinespace[1.5pt]
Fable 5 (medium; low) & HF & 28/48 (58.3\%) & 108.7$\rightarrow$92.9 & 18.0 & 23.4\%$\rightarrow$26.6\% & 4/40/3 \\
\addlinespace[1.5pt]
Grok 4.5 (high) & HF & 20/33 (60.6\%) & 141.5$\rightarrow$101.5 & 41.0 & 15.6\%$\rightarrow$18.8\% & 4/28/0 \\
\addlinespace[1.5pt]
Kimi K3 & HF & 15/31 (48.4\%) & 85.7$\rightarrow$72.9 & 25.0 & 21.9\%$\rightarrow$23.4\% & 1/27/2 \\
\addlinespace[1.5pt]
GPT-5.6 Luna (xhigh) & HF & 13/30 (43.3\%) & 175.5$\rightarrow$111.6 & 59.0 & 18.8\%$\rightarrow$18.8\% & 0/27/2 \\
\addlinespace[1.5pt]
DeepSeek V4 Pro & HF & 11/22 (50.0\%) & 86.0$\rightarrow$54.8 & 42.0 & 15.6\%$\rightarrow$18.8\% & 3/19/0 \\
 & 3-formula & 6/15 (40.0\%) & 44.3$\rightarrow$27.1 & 28.0 & 12.5\%$\rightarrow$15.6\% & 2/13/0 \\
\addlinespace[1.5pt]
Kimi K2.7 Code & HF & 8/22 (36.4\%) & 58.9$\rightarrow$49.9 & 12.0 & 18.8\%$\rightarrow$18.8\% & 0/17/4 \\
\addlinespace[1.5pt]
Gemini 3.5 Flash & HF & 2/18 (11.1\%) & 24.9$\rightarrow$24.2 & 6.5 & 20.3\%$\rightarrow$20.3\% & 0/18/0 \\
 & 3-formula & 2/14 (14.3\%) & 24.3$\rightarrow$23.1 & 8.5 & 15.6\%$\rightarrow$15.6\% & 0/14/0 \\
\addlinespace[1.5pt]
Kimi K2.6 & HF & 3/15 (20.0\%) & 25.4$\rightarrow$24.1 & 5.0 & 17.2\%$\rightarrow$17.2\% & 1/14/0 \\
 & 3-formula & 5/14 (35.7\%) & 39.7$\rightarrow$36.1 & 9.0 & 10.9\%$\rightarrow$10.9\% & 1/13/0 \\
\addlinespace[1.5pt]
Grok 4.3 & HF & 1/8 (12.5\%) & 18.9$\rightarrow$18.6 & 2.0 & 9.4\%$\rightarrow$9.4\% & 0/8/0 \\
\bottomrule
\end{tabular}
\caption{Per-configuration exact simplification on Challenge64, expanding panel (b) of Table~\ref{tab:supp-quality-generalization}. Inputs are final train-valid formulas from symbolic conditions; Smaller counts strict AST reductions that preserve every training prediction. AST means are over those inputs. H-exact uses the full task denominator, and world-exact counts compare the same generated holdout worlds before and after simplification.}
\label{tab:supp-challenge-config-simplification}
\end{table*}
}

\newcommand{\SupplementSymbolicFirstTable}{%
\begin{table*}[t]
\centering
\small
\setlength{\tabcolsep}{4pt}
\begin{tabular}{@{}lllrrrrr@{}}
\toprule
View & Front end & Downstream method & Final valid & H-exact & Residual valid & LLM calls/task & Calls saved \\
\midrule
B300 & \texttt{prenex} & HF & \SymbolicCascadeBenchmarkPrenexHFFinalValidPercent{}\% & \SymbolicCascadeBenchmarkPrenexHFHExactPercent{}\% & \SymbolicCascadeBenchmarkPrenexHFResidualValidPercent{}\% & \SymbolicCascadeBenchmarkPrenexHFCallsPerTask{} & \SymbolicCascadeBenchmarkPrenexHFCallsSavedPercent{}\% \\
B300 & \texttt{prenex} & Repeated & \SymbolicCascadeBenchmarkPrenexRepeatFinalValidPercent{}\% & \SymbolicCascadeBenchmarkPrenexRepeatHExactPercent{}\% & \SymbolicCascadeBenchmarkPrenexRepeatResidualValidPercent{}\% & \SymbolicCascadeBenchmarkPrenexRepeatCallsPerTask{} & \SymbolicCascadeBenchmarkPrenexRepeatCallsSavedPercent{}\% \\
B300 & \texttt{ad-mix} & HF & \SymbolicCascadeBenchmarkAdMixHFFinalValidPercent{}\% & \SymbolicCascadeBenchmarkAdMixHFHExactPercent{}\% & \SymbolicCascadeBenchmarkAdMixHFResidualValidPercent{}\% & \SymbolicCascadeBenchmarkAdMixHFCallsPerTask{} & \SymbolicCascadeBenchmarkAdMixHFCallsSavedPercent{}\% \\
B300 & \texttt{ad-mix} & Repeated & \SymbolicCascadeBenchmarkAdMixRepeatFinalValidPercent{}\% & \SymbolicCascadeBenchmarkAdMixRepeatHExactPercent{}\% & \SymbolicCascadeBenchmarkAdMixRepeatResidualValidPercent{}\% & \SymbolicCascadeBenchmarkAdMixRepeatCallsPerTask{} & \SymbolicCascadeBenchmarkAdMixRepeatCallsSavedPercent{}\% \\
\addlinespace[2pt]
C64 & \texttt{prenex} & HF & \SymbolicCascadeChallengePrenexHFFinalValidPercent{}\% & \SymbolicCascadeChallengePrenexHFHExactPercent{}\% & \SymbolicCascadeChallengePrenexHFResidualValidPercent{}\% & \SymbolicCascadeChallengePrenexHFCallsPerTask{} & \SymbolicCascadeChallengePrenexHFCallsSavedPercent{}\% \\
C64 & \texttt{prenex} & Repeated & \SymbolicCascadeChallengePrenexRepeatFinalValidPercent{}\% & \SymbolicCascadeChallengePrenexRepeatHExactPercent{}\% & \SymbolicCascadeChallengePrenexRepeatResidualValidPercent{}\% & \SymbolicCascadeChallengePrenexRepeatCallsPerTask{} & \SymbolicCascadeChallengePrenexRepeatCallsSavedPercent{}\% \\
C64 & \texttt{ad-mix} & HF & \SymbolicCascadeChallengeAdMixHFFinalValidPercent{}\% & \SymbolicCascadeChallengeAdMixHFHExactPercent{}\% & \SymbolicCascadeChallengeAdMixHFResidualValidPercent{}\% & \SymbolicCascadeChallengeAdMixHFCallsPerTask{} & \SymbolicCascadeChallengeAdMixHFCallsSavedPercent{}\% \\
C64 & \texttt{ad-mix} & Repeated & \SymbolicCascadeChallengeAdMixRepeatFinalValidPercent{}\% & \SymbolicCascadeChallengeAdMixRepeatHExactPercent{}\% & \SymbolicCascadeChallengeAdMixRepeatResidualValidPercent{}\% & \SymbolicCascadeChallengeAdMixRepeatCallsPerTask{} & \SymbolicCascadeChallengeAdMixRepeatCallsSavedPercent{}\% \\
\bottomrule
\end{tabular}
\caption{Symbolic-first results pooled over the matched configurations in each benchmark view. The indicated symbolic solver runs first; HF or repeated generation is applied only to the remaining unsolved tasks. Accepted Z3 formulas are evaluated as returned, without final exact simplification; routed HF solutions use their final simplified formulas. Final valid and H-exact use the full task denominator, whereas residual valid is restricted to tasks not solved by the symbolic front end. LLM calls/task is averaged over the original task set. We count at most one call per task and round; automatic API retries are excluded. Calls saved is measured against running the same downstream method without symbolic-first routing.}
\label{tab:supp-symbolic-first}
\end{table*}
}
\newcommand{\SupplementDifficultyRepairTable}{%
\begin{table*}[t]
\centering
\small
\setlength{\tabcolsep}{5pt}
\begin{tabular}{@{}llrrr@{}}
\toprule
\multicolumn{5}{l}{\textbf{(a) Difficulty before the methods diverge}}\\
View & Final outcome & $n$ & Median R1 error & Predicted repeat success \\
\midrule
B300 & Both valid & \SuppDifficultyBenchmarkBothN{} & \SuppDifficultyBenchmarkBothMismatchPercent{}\% & \SuppDifficultyBenchmarkBothRepeatSuccessPercent{}\% \\
B300 & HF only & \SuppDifficultyBenchmarkHFOnlyN{} & \SuppDifficultyBenchmarkHFOnlyMismatchPercent{}\% & \SuppDifficultyBenchmarkHFOnlyRepeatSuccessPercent{}\% \\
\addlinespace[1.5pt]
C64 & Both valid & \SuppDifficultyChallengeBothN{} & \SuppDifficultyChallengeBothMismatchPercent{}\% & \SuppDifficultyChallengeBothRepeatSuccessPercent{}\% \\
C64 & HF only & \SuppDifficultyChallengeHFOnlyN{} & \SuppDifficultyChallengeHFOnlyMismatchPercent{}\% & \SuppDifficultyChallengeHFOnlyRepeatSuccessPercent{}\% \\
\midrule
\multicolumn{5}{l}{\textbf{(b) Effect of parent-derived repair}}\\
View & Model & Pairs & Reduced error & Became valid \\
\midrule
B300 & Grok 4.3 & \SuppRepairRowAPairs{} & \SuppRepairRowAReducedPercent{}\% & \SuppRepairRowAExactPercent{}\% \\
\addlinespace[1.5pt]
C64 & Fable 5 (medium; low) & \SuppRepairRowBPairs{} & \SuppRepairRowBReducedPercent{}\% & \SuppRepairRowBExactPercent{}\% \\
C64 & GPT-5.6 Terra (xhigh) & \SuppRepairRowCPairs{} & \SuppRepairRowCReducedPercent{}\% & \SuppRepairRowCExactPercent{}\% \\
C64 & GPT-5.6 Sol (xhigh) & \SuppRepairRowDPairs{} & \SuppRepairRowDReducedPercent{}\% & \SuppRepairRowDExactPercent{}\% \\
C64 & GPT-5.6 Luna (xhigh) & \SuppRepairRowEPairs{} & \SuppRepairRowEReducedPercent{}\% & \SuppRepairRowEExactPercent{}\% \\
C64 & Grok 4.5 (high) & \SuppRepairRowFPairs{} & \SuppRepairRowFReducedPercent{}\% & \SuppRepairRowFExactPercent{}\% \\
C64 & Grok 4.3 & \SuppRepairRowGPairs{} & \SuppRepairRowGReducedPercent{}\% & \SuppRepairRowGExactPercent{}\% \\
\bottomrule
\end{tabular}
\caption{Difficulty of additional solutions and the effect of repair. Panel~(a) pools matched task--model outcomes within each benchmark. R1 error is the fraction of training objects misclassified by the shared Round~1 formula. Predicted repeat success is an out-of-fold estimate from task structure, planted-concept complexity, and shared Round~1 diagnostics; it is used only for retrospective analysis. Panel~(b) uses configurations with complete parent links. Each pair consists of one invalid direct proposal and its strongest unambiguous same-round parent-derived descendant. Reduced error includes descendants that become valid; became valid requires exact agreement with every training label.}
\label{tab:supp-difficulty-repair}
\end{table*}
}
\newcommand{\FrontierDynamicsFigure}{%
\begin{figure*}[t]
\centering
\includegraphics[width=\textwidth]{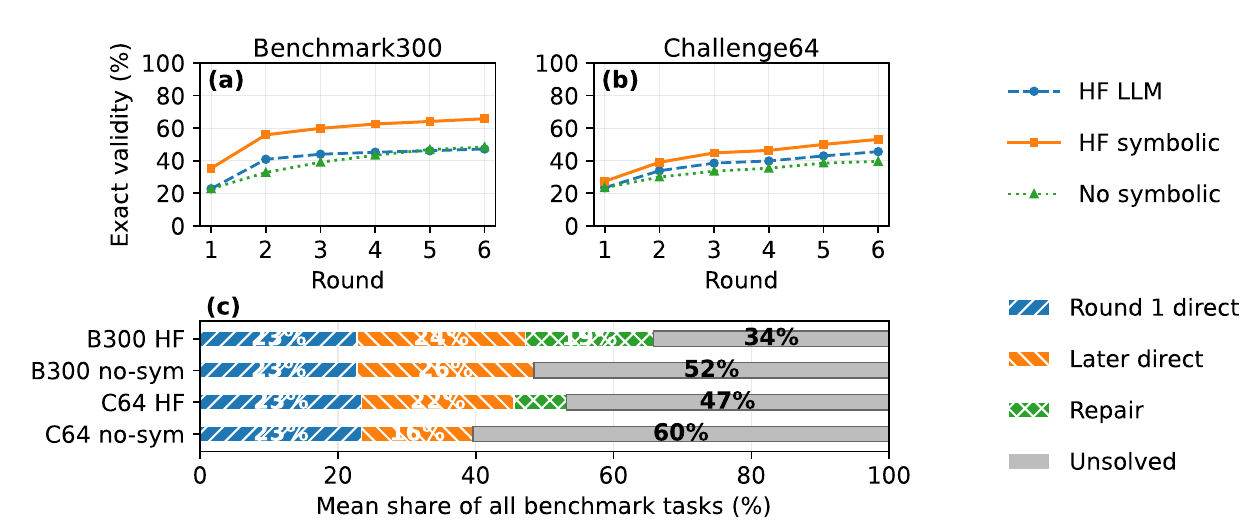}
\caption{Matched comparisons and first-source decomposition.
Panels (a) and (b) average cumulative exact validity over nine
matched single-formula/no-symbolic pairs: three on Benchmark300
and six on Challenge64. Dashed, solid, and dotted curves show
direct Hypothesis Frontier, post-symbolic Hypothesis Frontier,
and no-symbolic validity. Panel (c) assigns each valid formula to
its source, with the denominator being the full benchmark size.}
\label{fig:frontier-dynamics}
\end{figure*}%
}
\newcommand{\WorkedRepairView}{Benchmark300}
\newcommand{\WorkedRepairModel}{Grok 4.3}
\newcommand{\WorkedPartialProposalFormula}{\ensuremath{\displaystyle \exists y\,\bigl[S\bigl(x,y\bigr) \land P\bigl(y\bigr)\bigr]}}

\newcommand{\WorkedPartialProposalFP}{7}
\newcommand{\WorkedPartialProposalFN}{10}

\newcommand{\WorkedPartialRepairFormula}{\ensuremath{\displaystyle \exists y\,\bigl[S\bigl(x,y\bigr) \land \exists z\,\bigl[P\bigl(z\bigr) \land S\bigl(y,z\bigr)\bigr]\bigr]}}

\newcommand{\WorkedPartialRepairFP}{0}
\newcommand{\WorkedPartialRepairFN}{3}

\newcommand{\WorkedPartialLaterFormula}{\ensuremath{\displaystyle \exists y\,\bigl[S\bigl(x,y\bigr) \land \exists z\,\bigl[\bigl(P\bigl(z\bigr) \lor Q\bigl(z\bigr)\bigr) \land S\bigl(y,z\bigr)\bigr]\bigr]}}

\newcommand{\WorkedDirectProposalFormula}{\ensuremath{\displaystyle \exists y\,\bigl[S\bigl(x,y\bigr)\bigr]}}

\newcommand{\WorkedDirectProposalFP}{10}
\newcommand{\WorkedDirectProposalFN}{0}

\newcommand{\WorkedDirectRepairFormula}{\ensuremath{\displaystyle \exists y\,\bigl[S\bigl(x,y\bigr) \land \exists z\,\bigl[P\bigl(z\bigr) \land S\bigl(y,z\bigr)\bigr]\bigr]}}

\newcommand{\WorkedFactorPostFormula}{\ensuremath{\displaystyle \exists y\,\bigl[S\bigl(x,y\bigr) \land \exists z\,\bigl[S\bigl(y,z\bigr) \land \bigl(P\bigl(z\bigr) \lor Q\bigl(z\bigr)\bigr)\bigr]\bigr]}}
\newcommand{\WorkedFactorPreAST}{29}
\newcommand{\WorkedFactorPostAST}{17}

\newcommand{\WorkedHoldoutPostFormula}{\ensuremath{\displaystyle \exists y\,\bigl[P\bigl(y\bigr) \land \forall z\,\bigl[R\bigl(x,z\bigr) \lor \neg R\bigl(y,z\bigr)\bigr]\bigr]}}
\newcommand{\WorkedHoldoutPreAST}{38}
\newcommand{\WorkedHoldoutPostAST}{15}

\newcommand{\WorkedExtremePostFormula}{\ensuremath{\displaystyle \exists y\,\bigl[\forall z\,\bigl[S\bigl(x,z\bigr) \lor \neg R\bigl(y,z\bigr)\bigr] \land \neg Q\bigl(y\bigr)\bigr]}}
\newcommand{\WorkedExtremePreAST}{762}
\newcommand{\WorkedExtremePostAST}{16}

\newcommand{\WorkedExtremeBranchCount}{10}
\newcommand{\WorkedFactorPreLeftFormula}{\ensuremath{\displaystyle \exists y\,\bigl[S\bigl(x,y\bigr) \land \exists z\,\bigl[P\bigl(z\bigr) \land S\bigl(y,z\bigr)\bigr]\bigr]}}
\newcommand{\WorkedFactorPreRightFormula}{\ensuremath{\displaystyle \exists y\,\bigl[S\bigl(x,y\bigr) \land \exists z\,\bigl[Q\bigl(z\bigr) \land S\bigl(y,z\bigr)\bigr]\bigr]}}
\newcommand{\WorkedHoldoutPreFirstFormula}{\ensuremath{\displaystyle P\bigl(x\bigr)}}
\newcommand{\WorkedHoldoutPreCoreFormula}{\ensuremath{\displaystyle \exists y\,\bigl[P\bigl(y\bigr) \land R\bigl(x,y\bigr)\bigr]}}
\newcommand{\WorkedHoldoutPreGuardFormula}{\ensuremath{\displaystyle \forall z\,\bigl[R\bigl(x,z\bigr) \lor \neg S\bigl(z,x\bigr)\bigr]}}
\newcommand{\WorkedHoldoutPreExclusionFormula}{\ensuremath{\displaystyle \neg \exists w\,\bigl[R\bigl(x,w\bigr) \land \exists u\,\bigl[Q\bigl(u\bigr) \land R\bigl(u,w\bigr)\bigr]\bigr]}}

\newcommand{\WorkedPatternBenchmarkImmediatePercent}{29.6}
\newcommand{\WorkedPatternBenchmarkSolutionCount}{15}
\newcommand{\WorkedPatternBenchmarkImmediateSolutionCount}{6}

\newcommand{\WorkedPatternChallengeImmediatePercent}{30.0}
\newcommand{\WorkedPatternChallengeSolutionCount}{18}
\newcommand{\WorkedPatternChallengeImmediateSolutionCount}{5}

\title{Hypothesis Frontier: Verifier-Guided LLM--Symbolic Search
for First-Order Induction}

\author{Serafim Batzoglou}
\affiliations{Independent Researcher\\
\texttt{serafim.batzoglou@gmail.com}}

\begin{document}
\pagestyle{plain}

\maketitle
\thispagestyle{plain}

% BEGIN SILC-FOL REWRITE
\begin{abstract}
First-order concept synthesis asks a system to infer one
formula that classifies labeled objects consistently across
several finite relational structures. Every candidate can be
evaluated exactly, but quantified first-order formulas form a
vast search space, and LLM outputs are often semantically
promising without being fully correct. We introduce
\emph{Hypothesis Frontier}, a verifier-guided
neurosymbolic framework that evaluates each LLM formula
on every training object, retains the strongest verified
hypothesis across rounds, and uses its remaining errors to
guide subsequent generation. Symbolic processing repairs
invalid formulas while remaining anchored to the
LLM-generated hypothesis, and simplifies train-valid
formulas without changing any training prediction. Under
matched models, problem sets, and LLM-round budgets,
Hypothesis Frontier solves substantially more problems than
repeated original-prompt generation. After the final formulas
are selected, exact simplification shortens
many train-valid formulas while preserving every training
prediction; most behave the same on holdout worlds. Exact
symbolic reasoning therefore helps both to solve more induction
problems and to compress many of the resulting formulas.
\end{abstract}
% END SILC-FOL REWRITE

% ====================================================================
% GLOBAL AAAI MAIN-TEXT PLAN
% ====================================================================
% Abstract:             150--180 words
% Introduction:         600--700 words
% Problem Setting:      350--450 words
% Method:             1,250--1,450 words
% Experimental Design:  550--650 words
% Results:            1,250--1,500 words plus at most 3 empirical artifacts
% Related Work:         275--350 words
% Limitations:          180--230 words
% Conclusion:           120--160 words
%
% MAIN-TEXT HARD CONSTRAINT:
% All technical content must end by the bottom of page 7 in final AAAI layout.
% Pages 8--9 may contain references only.
%
% NARRATIVE:
% direct generation is brittle
% -> exact finite-world verification exposes actionable residuals
% -> symbolic editing and frontier prompting create persistent search
% -> the integrated loop beats repeated LLM generation
% -> LLM and symbolic moves contribute complementary progress
% -> compact selected hypotheses generalize much better.
%
% The abstract, numerical prose, and conclusion are finalized only after the
% audited result snapshot is frozen.
% ====================================================================

% BEGIN SILC-FOL REWRITE
\setcounter{secnumdepth}{1}
\section{Introduction}
\label{sec:introduction}

Inferring general rules from specific observations is a
fundamental form of learning. Logical induction makes those
rules explicit: given several finite relational
structures---small domains of objects with properties and
relations---and labels for an unknown concept, a system must
synthesize one first-order formula that matches every label
in every structure. The goal is easy to state and difficult to
achieve. Predicates, relational patterns, Boolean connectives,
and nested quantifiers generate a vast structured search
space, and correctness is exact: one misclassified object
invalidates the hypothesis.

LLMs and exact symbolic methods bring complementary
strengths. An LLM can propose coherent formulas combining
relations and quantifiers that bounded enumeration or local
rewriting may miss. Exact evaluation reveals every false
positive and false negative and verifies whether a revision
makes progress. Verifier-guided LLM systems often use such
checks to rank or reject completed outputs
\cite{cobbe2021gsm8k,polu2020generative}. We study a tighter integration in
which the strongest verified formula and its remaining errors
guide the next LLM call. Our question is whether this
recurrent neurosymbolic search is more reliable than LLM
generation alone.

We study the fully observed regime of the INDUCTION
benchmark \cite{induction2026}, a mechanically checkable
setting connecting relational induction
\cite{muggleton1991,quinlan1990} and solver-guided synthesis
\cite{solarlezama2008,alur2013sygus,torlak2014rosette}. Inputs are explicit relational
structures and outputs are executable first-order formulas, so
every candidate can be evaluated exactly. The observed
behavior is fully specified, but the explanatory rule remains
unknown.

We introduce \emph{Hypothesis Frontier}, which turns exact
feedback into iterative search. Each LLM formula is
evaluated on every training object. Invalid formulas are
repaired using the objects they misclassify; train-valid
formulas are simplified only when every training prediction
is preserved. A deterministic train-only ranking keeps the
strongest verified formula as the frontier, so progress is not
lost when a later output is worse. Symbolic search edits the
LLM formula or one of its descendants; it never substitutes an
independently synthesized formula. After the final frontier is
chosen, a final exact pass shortens train-valid formulas while
preserving their predictions on the training worlds.

An intermediate formula can help before it solves the
problem. A repair that removes some errors can replace its
parent formula as the frontier and be shown to the next LLM
call together with the remaining errors. Exact evaluation
therefore determines which formula is kept, how it is
changed, and where the next LLM round begins.

We evaluate Hypothesis Frontier in two FullObs views.
Benchmark300 spans several model families; Challenge64 is
a deliberately difficult subset evaluated with a different set
of models that includes several newer systems. Their
absolute success rates are therefore not directly comparable.
Across configurations with complete trajectories, mean
validity rises from
\FrontierInitialMeanBenchmarkPercent{}\% to
\FrontierFinalMeanBenchmarkPercent{}\% on Benchmark300
and from \FrontierInitialMeanChallengePercent{}\% to
\FrontierFinalMeanChallengePercent{}\% on Challenge64.
We compare Hypothesis Frontier with repeated LLM proposals under
the same LLM-round budget. Under matched models, problem sets, and
LLM-round limits,
Hypothesis Frontier wins \NoSymbolicWinCount{} of
\NoSymbolicPairCount{} comparisons, by
\NoSymbolicMinDeltaPoints{}--%
\NoSymbolicMaxDeltaPoints{} percentage points. The
problems solved only by Hypothesis Frontier start with
formulas farther from validity, and both later LLM proposals and
parent-derived repair contribute first exact solutions. The
final exact simplifier substantially reduces formula size and
modestly improves aggregate holdout validity, although most
formulas behave the same on holdout worlds before and after
simplification. Hypothesis Frontier therefore solves more
problems and shortens many train-valid formulas, while
recovering compact rules that remain correct on unseen worlds
is harder.

The \texttt{z3-prenex} + rescue solver and the specialized
\texttt{z3-ad-mix} portfolio \cite{induction2026} solve many
of the same problems as Hypothesis Frontier, and each
approach also solves problems the other misses. We therefore
test a symbolic-first workflow: run Z3 first, then apply
Hypothesis Frontier only to the unsolved problems. Across both
Z3 front ends, this workflow raises final validity by
\CascadeHFStandaloneGainMinPoints{}--%
\CascadeHFStandaloneGainMaxPoints{} percentage points
and reduces LLM calls by \CascadeHFCallSavingsMinPercent{}--%
\CascadeHFCallSavingsMaxPercent{}\% relative to running
Hypothesis Frontier on every problem.
On the same remaining problems, Hypothesis Frontier
exceeds repeated generation by
\CascadeResidualGainMinPoints{}--%
\CascadeResidualGainMaxPoints{} percentage points.
Together, these results show three roles for exact symbolic
reasoning: improving LLM-generated formulas during
search, simplifying exact formulas after search, and solving
some problems before any LLM call.

We make three contributions:
\begin{itemize}

\item \textbf{Verifier-guided search over LLM formulas.}
Hypothesis Frontier evaluates each formula exactly, repairs
its errors, keeps the strongest verified formula across calls,
and uses the remaining errors to guide the next call.
Symbolic search edits LLM-generated formulas rather than
replacing them with independently synthesized answers.

\item \textbf{Controlled comparisons and a symbolic-first
workflow.}
Under matched models, problem sets, and LLM-round limits,
Hypothesis Frontier outperforms repeated generation in
every comparison. Later LLM proposals and symbolic repair both
produce exact solutions, including on harder problems.
Running Z3 first solves additional problems and reduces LLM
calls; on the problems Z3 does not solve, Hypothesis
Frontier remains more effective than repeated generation.

\item \textbf{Exact simplification and concept recovery.}
After search ends, an exact simplifier shortens many
train-valid formulas while preserving every training prediction
and modestly improving overall holdout validity.
\end{itemize}
% END SILC-FOL REWRITE

% BEGIN SILC-FOL REWRITE
\section{Problem Setting}
\label{sec:problem-setting}

A FullObs instance from INDUCTION \cite{induction2026}
contains finite relational worlds
$\mathcal{W}=\{W_1,\ldots,W_m\}$ over the shared
signature $\Sigma=\{P,Q,R,S,=\}$, with unary predicates
$P,Q$ and binary predicates $R,S$. Each world $W$
provides a finite domain $D_W$, complete predicate
interpretations, and a target extension
$T_W\subseteq D_W$. Under the closed-world assumption,
every unlisted predicate fact is false. The task is to return
one first-order formula $\phi(x)$ with exactly one free
variable; the same formula is evaluated in every world.

The extension induced by $\phi$ in $W$ is
\[
\widehat{T}_W(\phi)
  = \{a\in D_W : W\models\phi[a]\}.
\]
The formula is \emph{train-valid} iff
$\widehat{T}_W(\phi)=T_W$ for every
$W\in\mathcal{W}$. Because the domains are finite, this
condition can be evaluated exactly on every labeled object. An invalid
formula therefore yields concrete false positives and false
negatives, not merely an aggregate error.

Exact agreement need not identify a unique concept. The
observed worlds may admit large case-splitting formulas
that exploit incidental finite-world structure. We therefore
report abstract syntax tree (AST) size alongside validity.
A planted reference formula
$\phi^\star$ serves only as a post-selection complexity
anchor; we do not assume that it is unique or globally
minimal.

At inference time, Hypothesis Frontier receives only the
observed worlds and target labels. The planted reference
formula and generated holdout worlds are used only after the
final reported formula is fixed, to evaluate compactness and
holdout exactness; they never affect prompting, repair,
simplification, ranking, retries, or stopping.
% END SILC-FOL REWRITE

% BEGIN SILC-FOL REWRITE
\section{Method: Hypothesis Frontier}
\label{sec:method}

\subsection{From Formula Generation to Verified Search}
\label{sec:method-overview}

Direct prompting treats each formula as final: execute it, then accept or discard it. This leaves useful information unused: an invalid formula may already express the right quantified relations while misclassifying only a few objects, and a train-valid formula may still contain unnecessary case structure. Hypothesis Frontier instead carries verified progress from one round to the next.

Each LLM proposal is parsed and evaluated on every training object. Unparseable outputs count as failures; parseable invalid formulas enter repair, train-valid formulas enter simplification, and every symbolic descendant is re-evaluated before retention. A deterministic train-only ranking then selects the cumulative frontier. The recurrent pipeline is LLM proposal \(\rightarrow\) exact verification \(\rightarrow\) repair or simplification \(\rightarrow\) frontier selection \(\rightarrow\) next proposal; the next prompt contains the selected frontier and any remaining errors.

All symbolic candidates are bounded, parent-derived edits of an LLM formula or one of its descendants; independent whole-formula synthesis is evaluated only in the separate symbolic-first workflow. The LLM can introduce or reorganize logical structure beyond the bounded edit neighborhoods, while symbolic processing makes locally verified corrections and removes unnecessary syntax. A partial repair can therefore advance the frontier before the task is solved, so later rounds continue from the improved state rather than restart.

\subsection{Residual-Guided Repair of Invalid Hypotheses}
\label{sec:repair}

A parseable but invalid formula may retain useful quantified
structure. \emph{Verified repair} preserves it and uses the
misclassified training objects to select bounded,
parent-derived edits. Let $T_W$ be the target extension in
training world $W$. The verifier returns
\[
\begin{aligned}
\mathrm{FP}(\phi)
  &= \{(W,a):W\models\phi[a],\,a\notin T_W\},\\
\mathrm{FN}(\phi)
  &= \{(W,a):W\not\models\phi[a],\,a\in T_W\},
\end{aligned}
\]
with
$m(\phi)=|\mathrm{FP}(\phi)|+|\mathrm{FN}(\phi)|$.
False positives call for restriction and false negatives for
expansion; their identities, not only $m(\phi)$, guide
candidate generation.

Repair combines three complementary candidate
generators. A structural beam applies Boolean normalization
and factoring, subtree deletion, quantifier-body pruning,
constraint relaxation, and guards around formulas or
existential witnesses. A selector generator scores compact
conditions and their negations on every training object.
Conditions that reject false positives while preserving
positive objects act as \emph{restrictors} $r(x)$; conditions
that cover false negatives while avoiding negative objects act
as \emph{expansions} $e(x)$, yielding
\[
\phi\land r,\qquad \phi\lor e,\qquad
(\phi\land r)\lor e,\qquad(\phi\lor e)\land r.
\]
When one condition cannot cover the residual, a multi-term
generator combines selectors into conjunctive or disjunctive
patches. Structural candidates edit the parent AST; patches
retain $\phi$ as a component. None replaces it with an
independently synthesized formula.

Every candidate is executed on all training objects. A
descendant is retained only if it reaches train validity, lowers
mismatch, or, at equal mismatch, is simpler. Retained
candidates enter the frontier ranking defined below.
A train-valid child proceeds to verified simplification;
otherwise, the strongest retained descendant may become the
frontier, with its recomputed residual shown in the next LLM
prompt.

\subsection{Verified Simplification of Valid Hypotheses}
\label{sec:simplification}

A formula can be valid on the training worlds but too complex.
Models often produce bloated formulas that consist of
case-splitting on specific conditions in each training world,
rather than simple concepts that capture the regularity of the
target concept. We therefore apply one procedure,
\emph{verified simplification}, only to train-valid formulas;
invalid formulas remain in residual-guided repair.

Let $\mathbf{p}_{\mathcal W}(\phi)$ be the vector of truth
values produced by $\phi$ over every training pair $(W,a)$,
and let
$c(\phi)=(\text{AST size},\text{quantifier depth},
\text{equality count})$. From a train-valid formula $\phi_0$,
verified simplification generates parent-derived AST edits:
Boolean deletion and factoring, quantifier pruning and merging,
equality simplification, and reuse of existing subformulas. A
candidate $\psi$ may replace the incumbent $\phi$ only if
\[
\mathbf{p}_{\mathcal W}(\psi)=
\mathbf{p}_{\mathcal W}(\phi_0)
\quad\text{and}\quad
c(\psi)<_{\mathrm{lex}}c(\phi).
\]
Exact execution checks the first condition on every training
object; the second makes accepted edits monotonically simpler.
Search resumes from the best verified descendants until no
candidate improves the incumbent or the budget is exhausted.

Across the final frontier of valid formulas, we apply a bounded
beam of exact simplification edits, including coordinated
Boolean and relation edits and the merging of compatible
quantified branches. Because this final pass runs after search
ends, it can change only the reported formula, not later
prompts, stopping, or LLM calls.

The guarantee is finite-world behavioral preservation, not
logical equivalence; holdouts assess behavior beyond the
training worlds.

\subsection{Frontier Selection and Subsequent Proposals}
\label{sec:frontier}

After each symbolic stage, all direct proposals and
parent-derived descendants seen through Round~$r$ are ranked
lexicographically. The order first prefers evaluable, then
parseable, then train-valid candidates; it next minimizes
mismatch count, AST size, quantifier depth, and equality
count. Source and round do not affect the ranking; a stable
identifier breaks final ties. The winner $F_r$ becomes the
cumulative frontier, preserving verified progress across calls.

Round~1 uses the original induction prompt to produce $H_0$.
Later \emph{single-best frontier prompts} repeat the full
problem, retain $H_0$ as an anchor, and supply $F_{r-1}$ with
its validity, residual counts, AST size, quantifier depth, and
misclassified objects. Round~2 uses the post-symbolic
Round~1 frontier; later rounds refresh this context after each
LLM and symbolic stage.

Each prompt requests one new formula, which re-enters
verify--repair--simplify before the expanded candidate bank is
reranked. Trajectories contain at most six LLM calls, one per
task and round, and may stop before Round~6 under the fixed
train-only policy; the last deterministic frontier is the final prediction.

Appendix Algorithms~\ref{alg:supp-repair}--%
\ref{alg:supp-frontier-update} provide pseudocode for
repair, verified simplification, and frontier updating;
Appendix Section~\ref{sec:supp-worked-trajectories} gives worked
trajectories.
% END SILC-FOL REWRITE

\section{Experimental Design}
\label{sec:experimental-design}

\subsection{Benchmarks and Comparisons}
\label{sec:benchmarks-systems}

We use Benchmark300 for broad comparisons across model
families. Challenge64 is a deliberately difficult subset used
to study newer models at manageable cost. The views use
different model sets, so we report them separately and do not
compare their absolute success rates.

Each configuration fixes the model and reasoning setting,
prompt, maximum number of LLM rounds, parser, symbolic search
procedure, ranking, and stopping rule. All symbolic
configurations use the same Round~1 prompt. Later Hypothesis
Frontier rounds request one formula from the current frontier;
Appendix Tables~\ref{tab:supp-benchmark-config-trajectories}--%
\ref{tab:supp-challenge-config-trajectories} report three-formula
variants that request three formulas in Rounds~2--6.
Table~\ref{tab:main-control-comparison} summarizes the evaluated
model--condition combinations and results; the cited sources
document the models
\cite{deepseek2026v4,google2026gemini35flash,kimi2026k26,kimi2026k27code,xai2025grok4,spacexai2026grok45,openai2026gpt56,anthropic2026fable5}.

We test whether symbolic search makes the same number of LLM
rounds more effective than repeated prompting. For each matched
model and task set, the no-symbolic pipeline repeats the
Round~1 prompt, uses the same round limit and train-only
ranking, and considers only formulas returned by the model. It
receives no frontier feedback and performs no repair or
simplification.

All matched conditions use the same verifier-independent
recovery procedure: one retry at the original reasoning setting,
one at the next lower setting, and a final lower-setting retry
only if the preceding pass recovered additional usable
responses. Tasks that still yield no usable formula count as
failures. Appendix Section~\ref{sec:supp-experimental-protocol}
gives the full recovery procedure and explains how calls are
counted. Each analysis
includes every configuration for which the required measurements
are available. Figure~\ref{fig:frontier-dynamics} uses all nine
matched configurations. The repair and formula-quality analyses
use seven single-formula configurations in which every symbolic
formula can be traced to its parent and no independently
synthesized formula enters the search. The configurations used
in the remaining analyses are listed with their results and in
Appendix Table~\ref{tab:supp-analysis-configurations}.

We also evaluate two LLM-free Z3 systems from INDUCTION
\cite{induction2026}. \texttt{z3-prenex} + rescue searches a
generic bounded prenex grammar; \texttt{z3-ad-mix} adds
FullObs-specific schemas. Each receives
\SymbolicBaselineTimeoutMinutes{} minutes per task, and a
timeout or no solution counts as failure. We use these systems
as reference points, not matched controls, because their
search spaces and compute differ from those of Hypothesis
Frontier. In the \emph{symbolic-first workflow}, a
train-valid Z3 formula ends the task. Otherwise, the
corresponding Hypothesis Frontier or repeated-generation
trajectory proceeds unchanged. The failed Z3 formula is never
shown to the model or added to its candidates. Reported LLM-call
savings exclude automatic API retries and Z3 computation, so
they do not measure total runtime or cost.

\subsection{Evaluation}
\label{sec:controls-protocol}

After at most six LLM calls, we report whether the final
selected formula matches every training label. The formula
must also parse and follow the output language. Missing,
empty, unparsable, or otherwise unusable outputs count as
failures, and all rates use the full task denominator. We also
report direct evaluability, mean calls, and round-level
validity for direct proposals, post-symbolic frontiers, and the
cumulative frontier.

We report AST size and its difference from the planted
reference. Acc@gold+10 and
Acc@gold+25 require train validity and no more extra AST nodes
than their named thresholds; bloat exceeds the latter
threshold. We evaluate fixed generated holdouts only after
inference. World exactness
means no errors in one holdout world. H-exact means no errors
in any holdout world and uses the full denominator. In
symbolic-first results, final validity and H-exact use all
task--model pairs, residual validity includes only Z3
failures, and calls/task is the number of reported LLM calls per
original pair.

For Hypothesis Frontier and the three-formula condition, the
reported train-valid formula is the output of the final exact
simplification pass defined above. Because this pass begins only
after the final train-only frontier is fixed, it cannot affect
prompting, calls, or stopping. Table~\ref{tab:main-control-comparison}
reports paired mean AST before and after the pass; if no smaller
formula is found, the original is retained. We evaluate Z3
formulas without this final pass. In the symbolic-first workflow,
tasks routed to Hypothesis Frontier use its final simplified
formula. No-symbolic rows report their final direct AST.

To understand why gains in train validity shrink under
stricter measures, we compare which tasks only one pipeline
solves, how difficult the tasks were before the pipelines
diverged, and whether shorter formulas perform better on
holdouts for the same task and model. We repeat the size
comparison after removing exact reference matches. All main
results use the deterministic final frontier.

\catcode95=11\relax
\catcode49=11\relax
% BEGIN SILC-FOL REWRITE
\section{Results}
\label{sec:results}

\subsection{How Does Symbolic Reasoning Improve
LLM-Based Induction?}
\label{sec:results-main}

Across matched model, task set, and LLM-round limits,
Hypothesis Frontier outperforms repeated original-prompt
generation in \NoSymbolicWinCount{} of
\NoSymbolicPairCount{} comparisons by
\NoSymbolicMinDeltaPoints{}--%
\NoSymbolicMaxDeltaPoints{} percentage points;
\NoSymbolicCIExcludesZeroCount{} paired bootstrap intervals
exclude zero (Table~\ref{tab:main-control-comparison}).
Repeated generation improves beyond Round~1 in
\NoSymbolicSamplingProgressCount{} runs, yet Hypothesis
Frontier uses fewer LLM calls in
\NoSymbolicLowerMeanCallsCount{} comparisons. Final H-exact
and Acc@gold+25 point estimates are higher for Hypothesis
Frontier in \PosthocHExactWinCount{} and
\PosthocAccGoldTwentyFiveWinCount{} of
\PosthocMatchedPairCount{} comparisons, respectively. On
Benchmark300, all \PosthocBenchmarkPairCount{} Acc gains range
from \PosthocBenchmarkAccGainMinPoints{} to
\PosthocBenchmarkAccGainMaxPoints{} points, and all
\PosthocBenchmarkAccCIExcludesZeroCount{} paired
\PosthocBootstrapConfidencePercent{}\% intervals exclude zero.
For each of the three model families evaluated in both benchmark
views, the one-formula variant reaches higher final validity than
the three-formula variant (Appendix
Tables~\ref{tab:supp-benchmark-config-trajectories}--%
\ref{tab:supp-challenge-config-trajectories}). Because
the variants also differ in their between-round symbolic
processing, we treat this as a comparison of complete pipelines
rather than a clean test of response width.

The additional solutions do not come only from cases already
close to a direct answer. Among configurations with the
measurements available before the two methods diverge, problems
solved only by Hypothesis Frontier have larger Round~1 mismatch
and a lower cross-fitted probability that repeated generation
will succeed. Thus the gain extends to problems that are harder
for the underlying model, although the analysis is observational.

\MainControlComparisonTable

Table~\ref{tab:main-control-comparison} separates the gain
from recurrent search from the effect of final simplification.
Valid and H-exact evaluate the final reported formulas,
whereas AST pre$\rightarrow$post compares each train-valid
formula before and after the final exact simplification pass.
The pass lowers mean AST in every symbolic configuration
without changing train validity, calls, or stopping.
No-symbolic rows report their direct AST because they receive
no simplification. We examine the aggregate simplification
and holdout results below.

Independent symbolic synthesis provides a complementary
comparison
(Table~\ref{tab:symbolic-only-baselines}, panels~a--b).
The Z3 formulas in panel~(a) are evaluated as returned and do
not receive the final exact simplification pass.
Among the two Z3 systems, the FullObs-specific
\texttt{z3-ad-mix} portfolio attains higher validity, smaller
valid formulas, and higher H-exact than generic prenex
search. Neither symbolic synthesis nor Hypothesis Frontier
subsumes the other: Hypothesis-Frontier-only task--model
cases outnumber Z3-only cases in every aggregate
comparison, while \texttt{z3-ad-mix} still solves cases that
Hypothesis Frontier misses. Most Benchmark300 gains over
repeated generation overlap with at least one Z3 system,
whereas most corresponding Challenge64 gains remain
unsolved by both. Fixed-language symbolic synthesis and
LLM-guided symbolic search therefore cover different
regions of the problem space.

Across both Z3 front ends, accepting train-valid Z3 outputs
before Hypothesis Frontier raises final validity by
\CascadeHFStandaloneGainMinPoints{}--%
\CascadeHFStandaloneGainMaxPoints{} percentage points
and reduces LLM calls by \CascadeHFCallSavingsMinPercent{}--%
\CascadeHFCallSavingsMaxPercent{}\% relative to running
Hypothesis Frontier on every task.
Table~\ref{tab:symbolic-only-baselines}(c) shows the prenex-first
cascade by configuration; Appendix
Table~\ref{tab:supp-symbolic-first} reports both Z3 front ends and
both downstream methods. On the same
residual cases, all \CascadeComparisonCount{}
comparisons favor Hypothesis Frontier over repeated
generation in final validity, residual validity, and LLM calls;
its residual validity advantage is
\CascadeResidualGainMinPoints{}--%
\CascadeResidualGainMaxPoints{} percentage points.
Symbolic reasoning thus helps at two levels: independent
synthesis solves some tasks before any model call, while
verifier-guided editing makes the remaining LLM search
more effective. We next examine where the additional exact
solutions enter the recurrent search.

\SymbolicOnlyBaselineTable

\subsection{When Do Additional Exact Solutions Emerge?}
\label{sec:results-frontier}

The additional solutions accumulate across several rounds.
Across the nine matched configurations in
Figure~\ref{fig:frontier-dynamics}, trajectories continue to solve
additional tasks after Round~1, with further progress in later
rounds. By the final
round, the post-symbolic frontier covers more exact solutions than
all direct LLM proposals combined in every matched configuration
(Figure~\ref{fig:frontier-dynamics}, panels~a--b).
Configuration-level trajectories for Benchmark300 and Challenge64
appear in Appendix
Tables~\ref{tab:supp-benchmark-config-trajectories}
and~\ref{tab:supp-challenge-config-trajectories}.

\FrontierDynamicsFigure

Within Hypothesis Frontier, both direct LLM generation and
parent-derived repair produce first exact solutions. Later
direct proposals are the largest direct source on Benchmark300,
whereas Round~1 and later direct proposals account for most
first solutions on Challenge64. The matched no-symbolic bars
contain no parent-derived repair by construction and leave a
larger unsolved share on both benchmarks
(Figure~\ref{fig:frontier-dynamics}, panel~c). This attribution
records where validity first appears: later direct proposals
receive a frontier shaped by the symbolic processing already
completed, so they are not an LLM-only counterfactual.

Repair often advances the frontier before it solves a
problem. Across the seven configurations with complete lineage records,
it lowers mismatch on
\RepairReducedMinPercent{}--\RepairReducedMaxPercent{}\%
of paired invalid proposals but reaches exact validity
immediately on only
\RepairExactMinPercent{}--\RepairExactMaxPercent{}\%.
Among the \PosthocRepairSelectedCount{} selected final
train-valid formulas produced by repair,
\PosthocRepairInvalidParentCount{}
(\PosthocRepairInvalidParentPercent{}\%) had train-invalid
parents. Because nearly every later call follows a partial repair,
these trajectories cannot show that repair caused the next proposal.
We treat mismatch reduction as verified search progress, not
evidence of a causal effect on the next model output.

\subsection{Simplifying Valid Formulas}
\label{sec:results-quality}

The final exact simplifier changes only train-valid formulas
and accepts a revision only when every training prediction is
preserved. It finds a smaller formula in
\SimplificationBenchmarkSmallerCount{} of
\SimplificationBenchmarkInputs{} cases
(\SimplificationBenchmarkSmallerPercent{}\%) on Benchmark300,
and \SimplificationChallengeSmallerCount{} of
\SimplificationChallengeInputs{} cases
(\SimplificationChallengeSmallerPercent{}\%) on Challenge64.
Mean AST falls from \SimplificationBenchmarkASTPreMean{} to
\SimplificationBenchmarkASTPostMean{} and from
\SimplificationChallengeASTPreMean{} to
\SimplificationChallengeASTPostMean{}, respectively, while
H-exact rises from
\SimplificationBenchmarkHExactPrePercent{}\% to
\SimplificationBenchmarkHExactPostPercent{}\% on Benchmark300
and from \SimplificationChallengeHExactPrePercent{}\% to
\SimplificationChallengeHExactPostPercent{}\% on Challenge64.
Paired world exactness usually remains unchanged.
Appendix Table~\ref{tab:supp-quality-generalization} reports the
aggregate analysis, and Appendix
Tables~\ref{tab:supp-benchmark-config-simplification}--%
\ref{tab:supp-challenge-config-simplification} report the results
for each configuration.

Formula size is most informative relative to the planted
reference. Every evaluated formula no larger than its
reference is holdout-exact in both benchmark views, whereas
the rate is at most \GoldBloatTaskExactMax{}\% for formulas
more than \ProtocolGoldLooseASTMargin{} nodes larger. The
reference is used only for this analysis, never during search
or simplification. Exact training fit alone is weaker evidence
of concept recovery: validity and H-exact are
\QualityBenchmarkValidPercent{}\% and
\QualityBenchmarkHExactPercent{}\% on Benchmark300, and
\QualityChallengeValidPercent{}\% and
\QualityChallengeHExactPercent{}\% on Challenge64.

Shorter formulas have a world-exactness advantage of
\CompactAllDelta_B{} percentage points on Benchmark300 and
\CompactAllDelta_C{} on Challenge64.
When both formulas remain larger than the reference, the
estimates are \CompactAboveGoldDelta_B{} points
[\CompactAboveGoldLo_B{}, \CompactAboveGoldHi_B{}] and
\CompactAboveGoldDelta_C{} points
[\CompactAboveGoldLo_C{}, \CompactAboveGoldHi_C{}]; neither
interval excludes zero. Thus simplification often produces a
smaller exact formula, but the strongest evidence of concept
recovery is reaching the scale of the planted reference.
% END SILC-FOL REWRITE

\catcode95=8\relax
\catcode49=\OtherCatcode\relax
% BEGIN SILC-FOL REWRITE
\section{Related Work}
\label{sec:related-work}

\textbf{Inductive logic programming and neural rule
induction.}
Inductive logic programming searches language-biased
hypothesis spaces for compact relational rules
\cite{quinlan1990,muggleton1991,cropper2022ilp}, while neural rule learners relax proof or rule
selection for gradient-based induction
\cite{evans2018dilp,rocktaschel2017proving}. Hypothesis Frontier retains
discrete first-order syntax and exact finite-model semantics.
INDUCTION introduced finite-structure concept synthesis and
reported both one-shot model behavior and bounded symbolic
synthesis \cite{induction2026}. The Z3 baselines make the
role of search bias explicit: one explores a generic prenex
grammar, while the other encodes FullObs-specific schemas.
Hypothesis Frontier takes a complementary route, using
LLM-generated formulas as evolving hypotheses that exact
symbolic search can repair, simplify, and retain across
rounds. The symbolic-first cascade instead tests whether an
independent solver can resolve some tasks before LLM inference.

\textbf{Program synthesis and execution-guided refinement.}
Sketching, counterexample-guided synthesis, and
syntax-guided synthesis alternate candidate construction
with formal verification \cite{solarlezama2008,jha2017synthesis,alur2013sygus,torlak2014rosette}.
Execution-guided LLM methods often rerank programs or
revise them using test outcomes, while Reflexion carries
linguistic
feedback across trials \cite{ni2023lever,chen2024selfdebug,shinn2023reflexion}.
Hypothesis Frontier combines exact object-level residuals
with parent-derived symbolic edits and a deterministic
frontier that conditions later proposals. Verification changes
the executable hypothesis and the next search state, not
merely a score or textual critique.

\textbf{Verifier-guided language models and logic benchmarks.}
External verifiers rank mathematical answers and guide
formal proof search \cite{cobbe2021gsm8k,polu2020generative,yang2023leandojo}. ProofWriter
and FOLIO primarily evaluate deduction from a supplied
theory \cite{tafjor2021proofwriter,han2024folio}, whereas Hypothesis Frontier
synthesizes an explicit first-order definition from labeled
finite worlds. In Hypothesis Frontier, exact verification transforms the
candidate formula and determines the next model context. Within
this loop, the LLM proposes candidate formulas and the symbolic
system verifies and revises them.
% END SILC-FOL REWRITE

% BEGIN SILC-FOL REWRITE
\section{Limitations}
\label{sec:limitations}

These results concern small, fully observed finite worlds over
a fixed vocabulary. This enables exact execution and
object-level residuals, but the evidence does not yet extend
to larger or partially observed structures, richer signatures,
theorem proving, or unrestricted first-order synthesis.
Hypothesis Frontier also trades computation for search through
multiple model proposals and verifier evaluations.

The no-symbolic comparison evaluates frontier conditioning,
repair, and simplification together rather than isolating them.
The Z3 systems use different search spaces and are not
compute-matched. The call-savings analysis excludes automatic
API retries and Z3 runtime, so it does not measure total
wall-clock time or cost. The mechanism and formula-quality
analyses are observational rather than causal. Generated
holdouts come from the same task generator, so they measure
generalization beyond the observed worlds without establishing
robustness to broader distribution shift. Finally, verified
simplification preserves behavior on the finite training worlds,
not logical equivalence; smaller formulas therefore need not
represent the same concept outside those worlds.

\par\noindent\textbf{AI assistance.} OpenAI Codex assisted with manuscript drafting and editing, code development,
data analysis, and figure preparation. The author reviewed and verified all
outputs and retains full responsibility for the work.
% END SILC-FOL REWRITE

% BEGIN SILC-FOL REWRITE
\section{Conclusion}
\label{sec:conclusion}

Hypothesis Frontier turns exact finite-model verification
from a final check into persistent LLM--symbolic search.
Under the same round budget, it solves substantially more
problems than repeated prompting, including problems whose
initial formulas are much farther from validity. The
trajectories show that later LLM proposals and
parent-derived repair both produce first exact solutions;
repair also often reduces errors without solving the problem
immediately.

The results reveal three roles for symbolic reasoning.
Independent synthesis can solve problems before any LLM
call, repair can develop an LLM-generated hypothesis during
search, and exact simplification can shorten the final valid
formula. Exact fit and shorter syntax, however, do not ensure
concept recovery; most simplified formulas behave similarly
on holdout worlds. The broader lesson is that an LLM
formula need not already be correct to be useful: with exact
feedback, it can become a hypothesis that symbolic methods
test, repair, retain, and simplify.
% END SILC-FOL REWRITE

\FloatBarrier
\bibliography{references}

\clearpage
\appendix
\setcounter{secnumdepth}{1}
\section*{Appendices: Extended Methods and Results}
\addcontentsline{toc}{section}{Appendices: Extended Methods and Results}

% BEGIN HYPOTHESIS FRONTIER SUPPLEMENT
\section{Overview}
\label{sec:supp-overview}

First-order concept synthesis makes inductive search directly observable:
every proposed formula can be executed on every labeled object, and every
retained symbolic change can be traced to its parent. The main text uses this
structure to study how exact verification, parent-derived repair, and later
LLM proposals work together. These appendices provide the experimental details,
per-configuration results, algorithms, and worked examples underlying those
analyses.

We begin with the evaluation protocol and the configurations used in each
analysis. We then present the full round-by-round, symbolic-first,
simplification, and holdout results; describe the repair, simplification, and
frontier procedures in more detail; and illustrate them with worked
trajectories. The final section explains how to verify the reported results
using the released code and data.

Throughout the appendices, train validity means exact agreement with every
label in the observed relational structures. Holdout exactness is evaluated
only after the final formula has been selected. Neither the planted reference
formula nor the generated holdout worlds influence prompting, repair,
simplification, ranking, retries, or stopping. Extended tables use the same
definitions and denominators as the main text.
% END HYPOTHESIS FRONTIER SUPPLEMENT

% BEGIN HYPOTHESIS FRONTIER SUPPLEMENT
\section{Experimental Protocol}
\label{sec:supp-experimental-protocol}

\subsection{Benchmarks and Validity}

Each task presents several fully observed finite structures with a shared
vocabulary and labels for an unknown unary concept. A system returns one
formula with free variable \(x\), and that formula is executed on every object
in every structure. The finite domains and complete predicate interpretations
make each prediction, error, and exact solution mechanically checkable.

Benchmark300 contains \ProtocolBenchmarkTaskCount{} tasks selected from the
\SupplementBenchmarkSourceTaskCount{} FullObs tasks. We used results from
\SupplementBenchmarkSelectionSystemCount{} systems to remove the simplest
tasks that were solved consistently, then used fixed random seed 20260530 to
subsample the next-easiest group and a subset of the hardest tasks that none
of the systems solved. The resulting benchmark is deliberately
difficulty-focused rather than a random sample, and the same
\ProtocolBenchmarkTaskCount{} tasks are used for every condition.

Challenge64 contains \ProtocolChallengeTaskCount{} Benchmark300 tasks. It
combines \SupplementChallengeCoreTaskCount{} difficult, discriminative tasks
with \SupplementChallengeExtensionTaskCount{} additional tasks selected
through a separate difficulty analysis. Because it uses a different set of
model configurations, its absolute rates should not be compared directly with
Benchmark300. Challenge64 is a computationally manageable stress test rather
than an unbiased estimate of Benchmark300 performance.

A task counts as solved only when the selected formula matches every training
label. Missing, malformed, and incorrect formulas remain failures. All validity
rates therefore use the full benchmark denominator rather than only tasks with
usable model outputs.

\subsection{Prompts and Compared Conditions}

The initial prompt contains the full training structures and labels, states the
closed-world semantics and allowed first-order grammar, and requests a
one-line JSON object containing one formula and a short description. Later
prompts in both Hypothesis Frontier variants retain the same induction problem
and add the Round~\ProtocolInitialRound{} formula, the current verified
frontier, and train-only verifier summaries: status, mismatch count, false
positives, false negatives, AST size, and quantifier depth. The one-formula
variant requests one formula in each later round, whereas the three-formula
variant requests three. No prompt contains the planted formula or holdout
worlds.

\begin{table*}[t]
\centering
\small
\setlength{\tabcolsep}{2.0pt}
\begin{tabular}{@{}lcccc@{}}
\toprule
Prompt content & Initial & \shortstack{One-formula\\later} &
\shortstack{Three-formula\\later} & \shortstack{Repeated\\later} \\
\midrule
Training structures and labels & yes & yes & yes & yes \\
Grammar and JSON contract & yes & yes & yes & yes \\
Verified formulas and errors & -- & yes & yes & -- \\
Requested formulas &
\ProtocolInitialPromptCandidateCount{} &
\ProtocolInitialPromptCandidateCount{} &
\ProtocolLaterPromptCandidateCount{} &
\ProtocolInitialPromptCandidateCount{} \\
\bottomrule
\end{tabular}
\caption{Information supplied to the model. HF denotes Hypothesis
Frontier. Its one- and three-formula variants receive the same categories of
training-side feedback shown here but request different numbers of formulas.
Repeated generation reissues the initial prompt.}
\label{tab:supp-prompt-contents}
\end{table*}

Each configuration fixes the model, reasoning setting, maximum number of LLM rounds,
parser, symbolic procedure, ranking rule, and stopping rule. Hypothesis
Frontier evaluates each response exactly, explores eligible parent-derived
edits, simplifies train-valid candidates, and carries the strongest verified
formula into the next call. The repeated-generation control uses the same
model, tasks, response-recovery schedule, LLM-round limit, and train-only
ranking. It repeats the initial prompt, admits only direct model formulas, and
receives no frontier feedback or symbolic processing.

Each reported model--condition--task cell comes from one executed trajectory;
we do not average repeated stochastic executions of the same cell. Confidence
intervals below resample the resulting task-level outcomes rather than
repeated model calls.

We selected the prompts, symbolic procedures, time limits, reasoning settings,
recovery rules, and the one- and three-formula variants using pilot
experiments and training-side diagnostics, before examining holdout results or
the complete benchmark outcomes.

For three model families, we also evaluate variants that request
\ProtocolLaterPromptCandidateCount{} formulas in each later round. Their
prompt format and between-round repair and simplification procedure differ
from those of the one-formula variants, so they compare complete pipelines
rather than isolating the effect of candidate count. Every LLM--symbolic
condition receives the same final exact simplification pass after search ends.

\subsection{Independent Symbolic Search}

We evaluate two LLM-free Z3 systems. \texttt{z3-prenex} + rescue enumerates
bounded quantifier prefixes and synthesizes a Boolean matrix over the available
atoms; its rescue schedule concentrates additional search on unsolved
quantified cases. \texttt{z3-ad-mix} is specialized to FullObs. It searches
compact schemas for universal-to-existential implications, guarded
existential--universal formulas, and two-hop existential chains, then falls
back to generic bounded templates. Both systems enforce every training label
as an exact SMT constraint and receive
\SymbolicBaselineTimeoutMinutes{} minutes per task.

The Z3 systems are independent reference points, not compute-matched controls.
In the symbolic-first workflow, a train-valid Z3 formula ends the task. If Z3
fails, the corresponding Hypothesis Frontier or repeated-generation method
runs exactly as it would without the Z3 attempt; the failed Z3 formula is not
shown to the model or added to the candidate set. The cascade therefore
measures the benefit of trying independent symbolic synthesis before either
LLM-based method.

\subsection{Response Recovery and Call Accounting}

Some API requests do not return a usable formula. Recovery follows the same
verifier-independent procedure in every matched condition: retry at the
original reasoning setting, then at the next lower setting, and make one final
lower-setting attempt only if the preceding pass recovered additional usable
responses. Fable 5 starts at medium reasoning and falls back to low because
higher settings rarely returned a usable formula.

Automatic API retries are not exposed consistently across providers. We
therefore count at most one LLM call per task and round, regardless of
automatic retries or the number of formulas recovered from a response. Each
trajectory has at most \ProtocolMaxLLMCalls{} reported calls and may stop
before that limit under the train-only stopping rule. We report direct evaluability
separately; symbolic descendants are not model calls.

\subsection{Computing Environment}

Model inference used the providers' APIs. Parsing, exact evaluation, symbolic
search, simplification, and analysis ran on an arm64 Mac
mini with an Apple M4 Pro (12 CPU cores and 16 integrated GPU cores), 24~GB
RAM, and macOS 15.6.1. The software environment used Python 3.12.2, Z3
4.15.4, NumPy 1.26.4, SciPy 1.13.1, pandas 2.2.2, scikit-learn 1.5.1,
Matplotlib 3.9.2, PyYAML 6.0.2, and ruamel.yaml 0.18.6. The local stages did
not require a discrete GPU.

\subsection{Post-Selection Evaluation}

Formula size is the number of AST nodes. Acc@gold+
\(\ProtocolGoldTightASTMargin{}\) and Acc@gold+
\(\ProtocolGoldLooseASTMargin{}\) require a train-valid formula to exceed the
planted reference by no more than the stated number of nodes. The planted
formula enters these retrospective measurements only.

We generated \SupplementHoldoutWorldsPerTask{} holdout worlds per task once
and saved them with the benchmark, using a deterministic task-specific seed
with offset \SupplementHoldoutSeedOffset{}. Their labels are computed from the
planted formula. Benchmark300 contains
\SupplementBenchmarkHoldoutGeneratedWorldCount{} of
\SupplementBenchmarkHoldoutRequestedWorldCount{} requested worlds, with at
least one world for \SupplementBenchmarkHoldoutTaskCoverage{} tasks.
Challenge64 contains \SupplementChallengeHoldoutGeneratedWorldCount{} of
\SupplementChallengeHoldoutRequestedWorldCount{} requested worlds, covering
\SupplementChallengeHoldoutTaskCoverage{} tasks. Tasks for which some holdout
worlds could not be generated remain in the evaluation; per-world validity
uses the number originally requested as its denominator.
H-exact requires correctness on every generated world for a task and uses the
full task denominator; a task with no generated world cannot count as exact.
Holdouts are consulted only after formula selection and never affect prompts,
repair, simplification, ranking, or stopping.

LLM--symbolic conditions report the final train-valid frontier after exact
simplification. This pass preserves every training prediction and begins only
after recurrent search ends, so it cannot change LLM calls or routing.
No-symbolic conditions report their selected direct formula unchanged. Z3
formulas in the symbolic-only and cascade analyses are also evaluated as
returned.

\subsection{Statistical Analysis}

For each matched model and benchmark, the validity difference is computed from
paired binary outcomes on the same tasks. Its
\SupplementConfidenceLevelPercent{}\% interval is the percentile interval from
\SupplementBootstrapResamples{} paired bootstrap resamples of task-level
differences, using fixed seed 271828. The bootstrap is applied separately to each matched
configuration; benchmark views are never pooled.

The difficulty analysis uses only quantities defined before the compared
methods diverge. A task-grouped
\SupplementCrossfitFolds{}-fold logistic model predicts repeated-generation
success from task structure, planted-concept complexity, and shared
Round~\ProtocolInitialRound{} response diagnostics. Formula-quality regressions
compare tasks solved by both methods with those solved only by Hypothesis
Frontier, controlling for normalized Round~\ProtocolInitialRound{} error,
planted AST and quantifier depth, training-set size, label prevalence, and
matched-configuration fixed effects; uncertainty is clustered by task. These
analyses test whether the additional tasks solved by Hypothesis Frontier were
already easier before the methods diverged. They do not affect the search and
do not establish that any individual symbolic edit caused a later success.
% END HYPOTHESIS FRONTIER SUPPLEMENT

% BEGIN HYPOTHESIS FRONTIER SUPPLEMENT
\section{Configurations Used in Each Analysis}
\label{sec:supp-analysis-configurations}

Different analyses require different measurements. The end-to-end comparison
needs matched Hypothesis Frontier and repeated-generation runs for the same
model and tasks. The repair analysis also needs a traceable parent for every
symbolic formula, while paired quality analyses require final formulas and
post-selection measurements from both methods. We therefore include every
configuration that provides the measurements needed for each question rather
than restricting all analyses to the smallest common subset.

Each model label corresponds to one fixed API model and reasoning setting; the
released prompts include the exact request parameters. Because reasoning
controls from different providers are not directly comparable, all
comparisons are made within the same model and task set rather than across
provider-specific effort labels.

The main comparison and every panel of the frontier figure use all
\NoSymbolicPairCount{} matched Hypothesis Frontier and repeated-generation
configurations. The separate parent-linked repair analysis uses
\FrontierConfigCount{} configurations whose symbolic descendants can be traced
to their parents, allowing each accepted edit to be compared directly with the
formula from which it was derived.

Comparisons of final formula quality on the same task and model use
\PairedQualityConfigCount{} matched configurations with complete outcomes and
measurements made before the methods diverge. The same set supports the overlap
comparison with independent Z3 synthesis. Table~\ref{tab:symbolic-only-baselines}(c) reports the
\PrenexCascadeConfigCount{} configurations for which the full prenex-first
cascade was evaluated. Appendix Table~\ref{tab:supp-symbolic-first}
reports both Z3 front ends and
both downstream methods for \TwoFrontEndCascadeConfigCount{} matched
configurations. Finally, the simplification analysis includes every
train-valid formula from the LLM--symbolic conditions reported across the main
text and appendices. It needs only the formula before and after
simplification, not its complete search history; no-symbolic formulas are
retained unchanged.

\SupplementAnalysisConfigurationTable
% END HYPOTHESIS FRONTIER SUPPLEMENT

% BEGIN HYPOTHESIS FRONTIER SUPPLEMENT
\section{Extended Results}
\label{sec:supp-extended-results}

The round-level results show how verified progress accumulates
(Table~\ref{tab:supp-rounds-simplification}). Each Hypothesis Frontier
entry reports cumulative train validity before and after symbolic processing.
The first value combines the incoming verified frontier with newly valid LLM
proposals from the current round; the second also includes the symbolic
descendants accepted in that round. Symbolic processing already raises validity
after the initial proposal and continues to contribute exact solutions later.

The pre-symbolic values inside Hypothesis Frontier are not an LLM-only
baseline. Each later proposal is conditioned on a frontier shaped by
verification and symbolic processing from completed rounds. Repeated
generation is therefore the relevant control: it holds the model, tasks, and
LLM-round limit fixed while removing frontier feedback and symbolic edits. The
separation grows over several rounds,
consistent with an advantage accumulated through recurrent search rather than
obtained from a single repair pass.

\SupplementRoundAndSimplificationTable

\subsection{Difficulty and Repair Progress}
\label{sec:supp-difficulty-repair}

The additional exact solutions found by Hypothesis Frontier are not simply
easy cases missed by repeated sampling
(Table~\ref{tab:supp-difficulty-repair}, panel~a). On Benchmark300, the median
Round~\ProtocolInitialRound{} error is
\SuppDifficultyBenchmarkHFOnlyMismatchPercent{}\% among task--model cases
solved only by Hypothesis Frontier, compared with
\SuppDifficultyBenchmarkBothMismatchPercent{}\% when both methods solve the
task. The corresponding Challenge64 medians are
\SuppDifficultyChallengeHFOnlyMismatchPercent{}\% and
\SuppDifficultyChallengeBothMismatchPercent{}\%. A cross-fitted model of
repeated-generation success gives the same ordering in both views. Hypothesis
Frontier therefore expands coverage into a harder part of the matched task
distribution, which helps explain why its newly found formulas are less often
compact or holdout-exact.

Panel~(b) shows what parent-derived repair contributes before a task is solved.
Across every configuration in which each symbolic descendant can be traced to
its parent, the strongest
same-round repair reduces the training error in
\RepairReducedMinPercent{}--\RepairReducedMaxPercent{}\% of matched
parent--descendant pairs, while only
\RepairExactMinPercent{}--\RepairExactMaxPercent{}\% become train-valid
immediately. Repair therefore usually improves a formula rather than solving
the task in one step. The improved formula can become the new frontier and
condition the next call, so edits that stop short of exact validity can still
advance the recurrent search.

\SupplementDifficultyRepairTable

Tables~\ref{tab:supp-benchmark-config-trajectories}
and~\ref{tab:supp-challenge-config-trajectories} report per-configuration
trajectories for Benchmark300 and Challenge64. They include every evaluated
Hypothesis Frontier, three-formula, and repeated-generation trajectory,
including Hypothesis Frontier runs without a matched control. The one- and
three-formula variants differ in both prompt format and between-round symbolic
processing. For the three model families shared across the two views, the
Challenge64 rows are the corresponding Benchmark300 runs evaluated on the
Challenge64 subset rather than separate model calls. The size of the symbolic
gain varies across models and rounds, but the frontier repeatedly preserves
and extends verified solutions.

\SupplementBenchmarkConfigurationTrajectoriesTable

\SupplementChallengeConfigurationTrajectoriesTable

Table~\ref{tab:supp-quality-generalization} reports the full formula-quality
analysis. Panel~(a) compares exact training fit with
compactness and holdout exactness. Panel~(b) isolates the final simplification
pass by pairing each train-valid input with its accepted output. Simplification
cannot create a new training solution or change an LLM call because it begins
only after the final train-valid formula has been selected. It nevertheless
finds a strictly smaller formula with identical training predictions in many
cases. Mean AST falls in both benchmark views, while most formulas retain the
same behavior on generated holdouts and H-exact changes more modestly.

\SupplementFormulaQualityTable

Tables~\ref{tab:supp-benchmark-config-simplification}
and~\ref{tab:supp-challenge-config-simplification} expand panel~(b) by
configuration. They show the number of formulas shortened, the change in mean
AST and H-exact, and the direction of each paired holdout change. The reduction
is not confined to one model or prompt condition: every reported symbolic
configuration has a lower mean AST after the pass. The size of the reduction
varies, as does its effect on unseen worlds.

\SupplementBenchmarkConfigurationSimplificationTable

\SupplementChallengeConfigurationSimplificationTable

Table~\ref{tab:supp-symbolic-first} shows a second role for exact symbolic
reasoning. Both Z3 front ends solve some tasks before any model call and reduce
the number of LLM calls over the full benchmark. Accepted Z3 formulas are
evaluated as returned, without the final
exact simplification pass. For tasks routed to Hypothesis Frontier, formula
quality is evaluated after that pass; routing and call counts remain unchanged.
The specialized \texttt{ad-mix} solver produces higher final
validity and larger call savings than generic prenex search in each benchmark
view. Its residual validity is not uniformly higher, because a stronger front
end removes more tasks before the LLM stage and leaves a smaller, more difficult
remainder.

For every benchmark view and symbolic front end, applying Hypothesis Frontier
to that remainder yields higher final and residual validity than repeated
generation. The two symbolic roles are therefore complementary: independent
synthesis removes tasks that need no model inference, while verifier-guided
search remains the stronger procedure on tasks that reach the LLM stage.

\SupplementSymbolicFirstTable

\subsection{Holdout Behavior and Formula Size}
\label{sec:supp-holdout-complexity}

Generated holdouts test whether a formula that is exact on the observed worlds
continues to recover the planted concept on fresh worlds from the same task
generator. Formula size is strongly associated with this outcome, but the
association is most informative relative to the planted reference
(Table~\ref{tab:supp-quality-generalization}, panel~c). Across both benchmark views,
every evaluated final formula no larger than its reference is holdout-exact;
the minimum rate across the two views is
\GoldOrSmallerTaskExactMin{}\%. When a formula exceeds the reference by more
than the stated margin, the corresponding rate is at most
\GoldBloatTaskExactMax{}\%. The reference is used only for this retrospective
comparison and never enters search, simplification, or frontier selection.

This pattern does not imply that arbitrary reductions in AST size improve
generalization. Same-task, same-model comparisons initially favor the shorter
formula (Table~\ref{tab:supp-quality-generalization}, panel~d), but much of that
advantage comes from pairs in which the shorter formula exactly recovers the
planted reference. Removing those pairs substantially reduces the difference.
When both formulas remain larger than the reference, neither benchmark shows a
reliable advantage for the shorter candidate: both confidence intervals include
zero. Compactness is therefore most persuasive as evidence of concept recovery
when search reaches a reference-scale representation, not as a general rule
that a smaller oversized formula will behave better on new worlds.
% END HYPOTHESIS FRONTIER SUPPLEMENT

% BEGIN HYPOTHESIS FRONTIER SUPPLEMENT
\section{Additional Method Detail}
\label{sec:supp-method-detail}

\subsection{Exact Evaluation and Residuals}

Each proposed formula is first parsed into an abstract syntax tree and checked
against the task language. A well-formed candidate must use the permitted
predicates and have exactly one free variable, $x$. The evaluator then assigns
$x$ to every object in every training world and computes the truth value of the
formula under the complete interpretation of that world. This produces a
behavior vector
\[
  \mathbf{p}_{\mathcal W}(\phi)
  = \bigl(\mathbb{1}[W\models\phi[a]]\bigr)_{(W,a)},
\]
which records every training prediction made by $\phi$.

Comparison with the target labels yields two object-level residuals. A false
positive is an object selected by the formula but excluded from the target;
a false negative is a target object omitted by the formula. We retain the
identities of these objects and their worlds, rather than only the total number
of errors. The count determines whether a candidate has improved, while the
objects themselves reveal where a restriction or expansion must act.
We write
$m(\phi)=|\mathrm{FP}(\phi)|+|\mathrm{FN}(\phi)|$ for this mismatch count.
For executable formulas, the complexity tuple
$c(\phi)$ consists of AST size, quantifier depth, and equality count, compared
lexicographically in that order.

\subsection{Bounded Repair from an LLM Hypothesis}

Repair begins from a parseable LLM formula or one of its symbolic descendants.
It never discards that formula and substitutes an independently synthesized
answer. Instead, it explores bounded changes to the parent. Structural edits
normalize or factor Boolean expressions, remove subtrees, prune quantified
bodies, and relax constraints. An \emph{existential witness} is the object
assigned to a variable introduced by an existential quantifier. A \emph{guard}
is a condition attached to the parent formula or to such a witness to restrict
or expand when that part of the formula applies. These edits are useful when
the parent contains the right relational idea but expresses it with an
unnecessary branch or an overly strong condition.

The residual also supports targeted patches. A \emph{selector} is a compact
formula drawn from a fixed library of unary tests and short quantified
relation patterns, together with their negations. Each selector is executed on
the training worlds and ranked by the errors it removes and introduces. A
\emph{patch} attaches one or more selectors while retaining the parent formula
as a component. A selector $r(x)$ is a useful restrictor when it rejects false
positives while preserving positive objects; a selector $e(x)$ is a useful
expansion when it covers false negatives while avoiding negative objects. The
repair search tests forms such as
\[
  \phi\land r,\qquad \phi\lor e,\qquad
  (\phi\land r)\lor e,\qquad (\phi\lor e)\land r,
\]
and may combine several selectors when no single one covers the residual. A
\emph{negated-selector patch} uses the complement of a selector; a
\emph{multi-term patch} joins several selectors by conjunction or disjunction
to cover different errors. Every patch therefore remains anchored to an LLM
proposal.

No edit is accepted on syntactic plausibility alone. Each child is executed on
all training objects. It remains eligible only if it becomes train-valid,
reduces the number of errors, or preserves that number with a simpler formula.
The residual is then recomputed from the child. Consequently, a partial repair
can become the next frontier, but only after exact evaluation has established
that it improves the train-only ordering.

Algorithm~\ref{alg:supp-repair} summarizes the repair search. Its structural
beam may retain a parseable intermediate formula long enough to expose a useful
second edit. Such an intermediate does not leave the procedure unless the
resulting descendant improves on the original parent. The selector searches
begin from that same parent and broaden the candidate set when ordinary local
edits do not cover the residual. Generated formulas must remain within fixed
AST-size and quantifier-depth limits. Within repair, ``highest-ranked'' first
prefers train validity, then smaller mismatch, then lower $c(\phi)$. The
\emph{structural beam} is the bounded set $\mathcal B$ of best parseable
formulas retained between edit steps.

\subsection{Simplification and Frontier Invariants}

Once a formula is train-valid, repair gives way to verified simplification.
The simplifier uses Boolean deletion and factoring, quantifier pruning and
merging, equality simplification, and reuse of compatible subformulas. A
candidate may replace its parent only when it produces exactly the same
behavior vector on the training worlds and is smaller under the lexicographic
complexity measure: AST size, then quantifier depth, then equality count. The
same rule is used at two points. A bounded search supplies compact formulas
during recurrent search; after the final frontier is fixed, a broader search
uses more edits and restarts. The latter can shorten the reported formula, but
it cannot retroactively change a prompt, model call, or stopping decision.

Algorithm~\ref{alg:supp-simplification} gives the expanded schedule used for
the reported final formulas. Here
$\textsc{Adopt}(\phi,\psi)$ returns $\psi$ only when exact execution
confirms that it is parseable, train-valid, behaviorally identical to $\phi$ on
the training worlds, and strictly simpler; otherwise it returns $\phi$. The
relation-aware phase exposes equivalent forms and searches bounded
combinations of Boolean and relation edits. The witness phases merge compatible
quantified branches around syntax already present in the formula. Local descent
then evaluates individual edits and repeatedly accepts the best verified
reduction. The smaller recurrent-search schedule uses the same acceptance rule
with fewer edit families and restarts.

\begin{algorithm}[t]
\caption{Parent-derived repair of an invalid formula}
\label{alg:supp-repair}
\begin{algorithmic}[1]
\REQUIRE Parseable invalid formula $\phi_0$, training worlds $\mathcal W$,
labels $\mathbf y$, and bounded search budgets
\ENSURE Highest-ranked verified descendant, or $\phi_0$ if none improves it
\STATE Execute $\phi_0$ to obtain $\mathrm{FP}_0$, $\mathrm{FN}_0$, and
$m(\phi_0)$
\STATE $\mathcal C\gets\{\phi_0\}$; $\mathcal B\gets\{\phi_0\}$
\WHILE{the structural beam has budget}
  \STATE $\mathcal D\gets$ local edits of formulas in $\mathcal B$
  \STATE Add guards and patches selected using $\mathrm{FP}_0$ and $\mathrm{FN}_0$
  \STATE Execute every formula in $\mathcal D$; add the results to $\mathcal C$
  \STATE $\mathcal B'\gets$ best bounded set from $\mathcal B\cup\mathcal D$
  \IF{$\mathcal B'=\mathcal B$}
    \STATE \textbf{break}
  \ENDIF
  \STATE $\mathcal B\gets\mathcal B'$
\ENDWHILE
\STATE Rank compact selectors by the errors they remove and introduce
\STATE $\mathcal P\gets$ bounded negated-selector and multi-term patches of
$\phi_0$
\STATE Execute $\mathcal P$; add the results to $\mathcal C$
\STATE Discard formulas outside the syntax and complexity limits
\STATE $\mathcal A\gets\{\psi\in\mathcal C:\psi$ is train-valid, or
$m(\psi)<m(\phi_0)$, or $m(\psi)=m(\phi_0)$ and $c(\psi)<c(\phi_0)\}$
\RETURN the highest-ranked formula in $\{\phi_0\}\cup\mathcal A$
\end{algorithmic}
\end{algorithm}

\begin{algorithm}[t]
\caption{Verified simplification of a train-valid formula}
\label{alg:supp-simplification}
\begin{algorithmic}[1]
\REQUIRE Train-valid formula $\phi_0$, training worlds $\mathcal W$, labels
$\mathbf y$, and a simplification schedule
\ENSURE A train-equivalent formula no more complex than $\phi_0$
\STATE Canonicalize and execute $\phi_0$; \textbf{return} it if verification fails
\STATE $\phi\gets\phi_0$
\STATE $\phi\gets\textsc{Adopt}(\phi,\textsc{RelationEdits}(\phi))$
\STATE $\phi\gets\textsc{Adopt}(\phi,\textsc{WitnessCompose}(\phi))$
\STATE $\phi\gets\textsc{Adopt}(\phi,\textsc{ExpandedWitnessCompose}(\phi))$
\STATE $\mathcal R\gets$ common restart schedule
\IF{$\phi_0$ is above the size threshold}
  \STATE Append the additional restart schedule to $\mathcal R$
\ENDIF
\FORALL{restart budgets $b\in\mathcal R$}
  \REPEAT
    \STATE $\mathcal D\gets$ bounded local edits of $\phi$
    \STATE Execute $\mathcal D$; keep the train-valid formulas simpler than $\phi$
    \IF{no verified reduction remains}
      \STATE \textbf{break}
    \ENDIF
    \STATE $\phi\gets$ the simplest retained formula
  \UNTIL{budget $b$ is exhausted}
\ENDFOR
\STATE Re-execute $\phi$; \textbf{return} $\phi_0$ if final verification fails
\RETURN $\phi$
\end{algorithmic}
\end{algorithm}

\noindent\textbf{Simplification operators.}
The simplification schedule specifies which phases run and the time and
evaluation budget assigned to each phase. An \emph{atom} is one predicate
application, such as $P(x)$ or $R(x,y)$.
\textsc{RelationEdits} first moves negations to atoms through
equivalence-preserving rewrites. It then interleaves local descent with a
bounded exact search over Boolean child deletion and local changes to
connectives and quantifiers, predicate names, whether an atom is negated, and
the order of relation arguments, such as $R(x,y)$ versus $R(y,x)$. This exact
search jointly selects a bounded
combination of edits while enforcing every training label. Only a smaller
train-valid result is returned.

\textsc{WitnessCompose} finds compatible existential and universal branches,
prunes their bodies, and places the universal condition under an existential
witness already present in the formula. It may reuse compatible relation atoms
from the existential branch. \textsc{ExpandedWitnessCompose} additionally
allows composition within a shared Boolean branch and bounded substitution of
the relation predicate, such as replacing $R$ by $S$. Both procedures execute
every proposed composition and return only the best verified reduction.

A \emph{local edit} deletes or contracts Boolean branches, factors common
structure, removes Boolean terms implied by others, prunes or merges
quantifiers, removes vacuous binders, or reuses a smaller existing subformula. A
\emph{restart} reruns this deterministic local descent from the current formula
with a fresh evaluation and time budget. Every formula receives the common
restart schedule; formulas whose initial size exceeds a threshold fixed before
evaluation receive the additional schedule. Canonicalization only normalizes
the parsed representation before these verified edits.

After each round, the frontier is selected from every direct proposal and
every retained symbolic descendant seen so far. Unusable outputs rank below
parseable formulas. Among formulas that can be executed, the order first
prefers train validity, then fewer errors, then lower complexity; a stable
identifier resolves any remaining tie. Neither source nor round receives
preference. Because retained candidates stay in the pool, a new frontier
cannot rank below the incumbent under this order.

Algorithm~\ref{alg:supp-frontier-update} puts these steps together. The
repair and simplification procedures perform their own bounded searches and
return only verified results. If repair reaches train validity, simplification
is applied before the candidate pool is reranked. A rejected edit never enters
the cumulative pool.

% A single-column float prevents a nearly empty float-only appendix page.
\begin{algorithm}[t]
\caption{One verified frontier update}
\label{alg:supp-frontier-update}
\begin{algorithmic}[1]
\REQUIRE Training worlds $\mathcal W$, labels $\mathbf y$, incumbent candidate
pool $\mathcal C_{r-1}$, and direct LLM proposals $\mathcal H_r$
\ENSURE Frontier $F_r$ and its verified residual
\STATE $\mathcal C \gets \mathcal C_{r-1}\cup\mathcal H_r$
\FORALL{$h\in\mathcal H_r$}
  \IF{$h$ parses and can be executed on $\mathcal W$}
    \STATE Compute $\mathbf p_{\mathcal W}(h)$, its residual, and $m(h)$
    \STATE $\psi\gets h$
    \IF{$m(h)>0$}
      \STATE $\psi\gets\textsc{Repair}(h,\mathrm{FP}(h),\mathrm{FN}(h))$
    \ENDIF
    \IF{$\psi$ is train-valid}
      \STATE $\psi\gets\textsc{Simplify}(\psi)$ using the recurrent schedule
    \ENDIF
    \STATE Add $\psi$ to $\mathcal C$
  \ENDIF
\ENDFOR
\STATE $F_r \gets$ highest-ranked formula in $\mathcal C$
\RETURN $F_r$ and its recomputed false positives and false negatives
\end{algorithmic}
\end{algorithm}

In Algorithm~\ref{alg:supp-frontier-update}, \textsc{Repair} denotes
Algorithm~\ref{alg:supp-repair}, and \textsc{Simplify} denotes
Algorithm~\ref{alg:supp-simplification} with the smaller recurrent schedule.

These rules make the state carried between calls precise. Every symbolic
formula has a traceable LLM ancestor; every accepted repair has a verified
train-only improvement; every accepted simplification preserves all training
predictions; and frontier quality is cumulative. The planted reference and
holdout worlds play no role in any of these decisions.
% END HYPOTHESIS FRONTIER SUPPLEMENT

% BEGIN HYPOTHESIS FRONTIER SUPPLEMENT
\section{Worked Search Trajectories}
\label{sec:supp-worked-trajectories}

The examples below were chosen deliberately because each isolates a different
part of the method. They are illustrations, not estimates of average behavior;
the aggregate results appear in the preceding sections.

\subsection{A Repair That Advances the Frontier}

Consider one \WorkedRepairView{} trajectory under \WorkedRepairModel{}. The
initial proposal identifies a one-step relational pattern:
\[
  \WorkedPartialProposalFormula .
\]
It produces \WorkedPartialProposalFP{} false positives and
\WorkedPartialProposalFN{} false negatives. Repair replaces the unary
condition on the immediate neighbor by a second relational step:
\[
  \WorkedPartialRepairFormula .
\]
The repaired formula is still invalid, but its residual falls to
\WorkedPartialRepairFP{} false positives and \WorkedPartialRepairFN{} false
negatives. It therefore becomes the verified frontier. The next prompt receives
this formula and its remaining errors. Its direct proposal broadens the final
unary condition from $P$ to $P\lor Q$:
\[
  \WorkedPartialLaterFormula .
\]
Exact evaluation now finds no false positives or false negatives. The example
shows why an invalid repair can be useful: it converts a rough relational idea
into a stronger state from which the next round continues. It does not, by
itself, show that repair caused the following proposal to succeed.

Repair can also finish the task without another model call. In a second
trajectory, the direct formula
\(
  \WorkedDirectProposalFormula
\)
has \WorkedDirectProposalFP{} false positives and
\WorkedDirectProposalFN{} false negatives. A parent-derived edit adds the
missing second step,
\(
  \WorkedDirectRepairFormula
\),
and the resulting formula is train-valid.

\subsection{How Often Does Repair Precede a Later Solution?}

Across all eligible trajectories, a partial repair immediately precedes
\WorkedPatternBenchmarkImmediateSolutionCount{} of
\WorkedPatternBenchmarkSolutionCount{} later-round first solutions on
Benchmark300 and \WorkedPatternChallengeImmediateSolutionCount{} of
\WorkedPatternChallengeSolutionCount{} on Challenge64. However, partial repair
also precedes \WorkedPatternBenchmarkImmediatePercent{}\% and
\WorkedPatternChallengeImmediatePercent{}\% of all eligible later calls,
respectively. After adjustment, the confidence interval for its association
with next-call validity includes zero in both views. These results show that
partial repairs persist into later search, but not that they cause the next
model proposal to succeed.

\subsection{Three Forms of Exact Simplification}

The first simplification example is an ordinary factorization. Two disjuncts
repeat the same relational path and differ only in the final unary predicate:
{\small
\[
\begin{aligned}
  &\WorkedFactorPreLeftFormula \\
  &\quad\lor\ \WorkedFactorPreRightFormula \\
  &\qquad\longrightarrow\qquad \WorkedFactorPostFormula .
\end{aligned}
\]
}
The AST falls from \WorkedFactorPreAST{} to \WorkedFactorPostAST{} nodes. Both
formulas are correct on all four generated holdout worlds.
Here the shorter formula exposes the shared witnesses and replaces the two
branches by a single $P\lor Q$ condition.

Finite-world simplification can make a less obvious change because its
acceptance rule preserves predictions on the training structures rather than
requiring global logical equivalence. In a readable example, the pass changes
\[
\begin{aligned}
\WorkedHoldoutPreFirstFormula\ \lor\ \Bigl(&\WorkedHoldoutPreCoreFormula\ \land \\
 &\bigl(\WorkedHoldoutPreGuardFormula\ \lor \\
 &\quad\WorkedHoldoutPreExclusionFormula\bigr)\Bigr)
\end{aligned}
\]
to
\[
  \WorkedHoldoutPostFormula .
\]
The formulas make identical predictions on every training object, while the
AST falls from \WorkedHoldoutPreAST{} to \WorkedHoldoutPostAST{}. On the same
generated holdout worlds, exactness improves from one of five worlds to all
five. Holdout labels played no role in accepting the
edit. This case illustrates a possible benefit, not a guarantee: most paired
holdout outcomes in the aggregate analysis are unchanged.

An extreme case shows the reach of the same verified search. Its source is a
\WorkedExtremePreAST-node disjunction with
\WorkedExtremeBranchCount{} large top-level branches, schematically
\(\bigvee_i \psi_i\). The simplifier reduces it to
\[
  \WorkedExtremePostFormula ,
\]
with \WorkedExtremePostAST{} AST nodes and exactly the same training
predictions. Holdout exactness improves from zero of five worlds to all five.
This example is
intentionally exceptional. It demonstrates that a very long train-valid
formula can conceal a compact rule on the observed finite worlds; it does not
claim that the two formulas are logically equivalent beyond those worlds.
% END HYPOTHESIS FRONTIER SUPPLEMENT

% BEGIN HYPOTHESIS FRONTIER SUPPLEMENT
\section{Reproducibility}
\label{sec:supp-reproducibility}

The accompanying release contains the Benchmark300 and Challenge64 task
definitions, the generated holdout worlds, the aggregate results and formula
lineages used in the paper, the inputs used to produce the tables, and the
local parser and evaluator. It also provides file hashes and provenance
metadata for the principal empirical results.

From the release root, run:
\begin{verbatim}
python -m venv .venv
source .venv/bin/activate
pip install -e .
./scripts/verify_artifact.sh
\end{verbatim}
The verification script runs offline and requires no credentials. It checks
file integrity and table calculations, validates the provenance metadata, and
evaluates one Challenge64 task with the released evaluator. It does not rerun
model inference, symbolic search, or holdout generation. These materials are
sufficient to verify the reported results; reproducing the original API calls
would require access to the corresponding model services.
% END HYPOTHESIS FRONTIER SUPPLEMENT

\end{document}